\pdfoutput=1

\documentclass[11pt]{article}

\usepackage[final]{acl}

\usepackage{times}
\usepackage{latexsym}

\usepackage[T1]{fontenc}

\usepackage[utf8]{inputenc}

\usepackage{microtype}

\usepackage{inconsolata}

\usepackage{graphicx}

\usepackage{amsmath,amsfonts,bm}

\def\eqref#1{equation~\ref{#1}}

\def\1{\bm{1}}

\DeclareMathAlphabet{\mathsfit}{\encodingdefault}{\sfdefault}{m}{sl}
\SetMathAlphabet{\mathsfit}{bold}{\encodingdefault}{\sfdefault}{bx}{n}

\usepackage{booktabs}
\usepackage{xspace}

\usepackage{amsmath}

\usepackage{titlesec}
\titlespacing*{\paragraph}   {0pt}{.6ex plus .3ex minus .2ex}{1em}

\usepackage{svg}

\usepackage{microtype}

\usepackage[]{bbm}

\usepackage{inconsolata}

\usepackage{tikz}
\usetikzlibrary{intersections}
\usepackage{pgfplots}
\usepackage{pgfplotstable}
\pgfplotsset{compat=1.7}
\usepackage{subcaption}
\usepgfplotslibrary{groupplots}
\usepgfplotslibrary{fillbetween}
\usetikzlibrary{matrix}
\usetikzlibrary{positioning}
\usepackage{fnpct}

\definecolor{c0}{HTML}{1f77b4} %
\definecolor{c1}{HTML}{2ca02c} %
\definecolor{c2}{HTML}{ff7f0e} %
\definecolor{c3}{HTML}{9467bd} %
\definecolor{c4}{HTML}{d62728} %
\definecolor{c5}{HTML}{bcbd22} %
\definecolor{c6}{HTML}{8c564b} %
\definecolor{c7}{HTML}{7f7f7f} %
\definecolor{c8}{HTML}{17becf} %
\definecolor{c9}{HTML}{e377c2} %

\usepackage[normalem]{ulem}

\newcommand{\eat}[1]{}
\usepackage{lipsum}

\usepackage{titlecaps}
\Addlcwords{a an the in by via and from as of to for like over with that vs}
\let\OldParagraph\paragraph
\renewcommand{\paragraph}[1]{\OldParagraph{\titlecap{#1}}}

\newcommand{\aautoref}[1]{\hyperref[#1]{Appendix~\ref*{#1}}}

\newcommand{\nlstring}[1]{\textit{#1}}
\def\code#1{\texttt{#1}}

\newcommand{\reals}{\mathbb{R}}

\AtBeginDocument{
  \everydisplay{\small}
}

\newcommand{\multiref}{\textsc{MultiRef}\xspace}
\newcommand{\respect}{\textsc{ReSpect}\xspace}
\newcommand{\Idefics}{IDEFICS2-8B\xspace}
\newcommand{\kilogram}{\textsc{KiloGram}\xspace}
\newcommand{\feedbacktype}{\gamma}
\newcommand{\feedback}{\feedbacktype}

\newcommand{\action}{a}

\newcommand{\context}{x}

\newcommand{\utterance}{u}

\newcommand{\feedbackdecoder}{\phi}
\newcommand{\param}{\theta}
\newcommand{\round}{\rho}
\newcommand{\policy}{\pi}
\newcommand{\dataset}{D}

\newcommand{\futureutt}{\bar{f}}

\newcommand{\feedbackpos}{\texttt{positive}}
\newcommand{\feedbackneg}{\texttt{negative}}
\newcommand{\feedbackneu}{\texttt{neutral}}

\newcommand{\bp}{\textsc{b-fft}\xspace}
\newcommand{\ba}{\textsc{b-rl}\xspace}
\newcommand{\tp}{\textsc{t-fft}\xspace}
\newcommand{\ta}{\textsc{t-rl}\xspace}
\newcommand{\bk}{\textsc{b-kto}\xspace}
\newcommand{\tk}{\textsc{t-kto}\xspace}
\newcommand{\hh}{\textsc{hh}\xspace}
\newcommand{\control}{\textsc{control}\xspace}

\usepackage{tcolorbox}
\usepackage{enumitem}
\newcommand{\tkimg}{\code{<img>}}
\newcommand{\boxcomment}[1]{\textcolor{c0}{\emph{\dotfill(#1)}}}

\title{Retrospective Learning from Interactions}

\author{Zizhao Chen, Mustafa Omer Gul, Yiwei Chen, Gloria Geng, Anne Wu \& Yoav Artzi\\
Department of Computer Science and Cornell Tech, Cornell University\\
\texttt{\{czz,momergul,annewu,yoav\}@cs.cornell.edu} 
\texttt{\{yc833,gcg46\}@cornell.edu}
}

\begin{document}
\maketitle
\begin{abstract}
Multi-turn interactions between large language models (LLMs) and users naturally include implicit feedback signals. If an LLM responds in an unexpected way to an instruction, the user is likely to signal it by rephrasing the request, expressing frustration, or pivoting to an alternative task. Such signals are task-independent and occupy a relatively constrained subspace of language, allowing the LLM to identify them even if it fails on the actual task. We introduce \textsc{ReSpect}, a method to learn from such signals in past interactions via retrospection without additional annotations. 
We deploy \textsc{ReSpect} in a new multimodal interaction scenario, where humans instruct a multimodal LLM to solve an abstract reasoning task with a combinatorial solution space. Through thousands of interactions with humans, we show how \textsc{ReSpect} gradually improves task completion rate from 31\% to 82\%, all without any external annotation.

\end{abstract}

\section{Introduction}

\begin{figure*}[t]
\centering
\includegraphics[width=0.9\linewidth, trim={60 250 100 215}, clip]{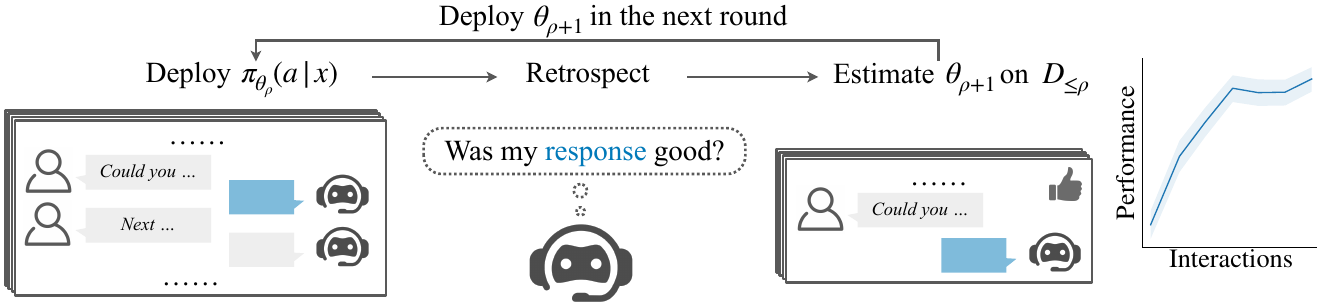}
\vspace{-5pt}
\caption{The \respect process. We deploy an MLLM policy $\policy_{\param_\round}(\action \vert \context)$ in rounds $\round$, to interact with users in multi-turn interactions. Following each round, the MLLM retrospectively analyzes each of its actions (highlighted in \textcolor{c0}{blue}) to decode feedback given the interaction context and  follow-up utterances. The decoded feedback can be positive (thumbs up as illustrated), negative or neutral. After each round, the model is retrained using all data aggregated so far $\dataset_{\leq\round}$. The MLLM improves over time without any external annotations. The plot on the right shows the performance curve in our experiments -- task success rate improves from 31\% to 82\% over six rounds.
}\label{fig:respect}
\vspace{-5pt}
\end{figure*}

Multi-turn interactions between human users and large language models (LLMs) are naturally rich with implicit learning cues. 
If the LLM fails to respond appropriately, the user may express frustration, rephrase their intent, or even completely pivot what they ask for. 
If the LLM does well, the user may express approval or simply continue to their next objective. 
This is not a property unique to LLMs, but a general characteristic of effective natural language communication~\cite{leavitt1951some}.
Such signals can inform the LLM of its performance, thereby creating an opportunity to learn through deployment, with no annotation cost.

We introduce \respect, a method for a model to learn from its own past interactions with human users. 
Rather than relying on feedback from annotators~\cite{ouyang2022Training} or assuming access to stronger models~\cite{bai2022Constitutional}, \respect relies solely on regular deployment interactions, where users interact with the model to achieve their goals. 
The key is using the deployment model (i.e., not a stronger or specialized model) to decode the implicit feedback expressed in follow-up human utterances in its own past interactions, thereby allowing the model to autonomously bootstrap from its interactions with human users.

We experiment with \respect by deploying a multimodal LLM (MLLM) in \multiref, a new multi-turn grounded interaction scenario (\autoref{sec:multiref}). \multiref is a challenging generalization of reference games~\citep{rosenberg1964Speakers} in that it requires humans to instruct the model step by step and models to reason about abstract visuals. 
While \respect is designed for broad and domain-independent deployment, \multiref provides an ideal test bed to answer our research questions in a lab environment. 
It naturally elicits the kind of gradual interactions common in conversational interactions with LLMs. 
It poses a challenging task to contemporary MLLMs, allowing us to observe a significant learning effect. At the same time, it is scoped, enabling a strong effect with limited data.

The key insight underlying \respect is that conversational implicit feedback signals occupy a relatively constrained subspace of natural language.
Such signals can include direct approvals (e.g., \nlstring{great!}) or signs of frustration (e.g., \nlstring{not again}), and also more subtle cues, such as when the user rephrases their request. 
Critically, it is relatively simple to disentangle them from task performance. 
A human can easily figure out from such cues if they do well or not, even if they have little understanding about what they are asked for.
It is this constrained nature that makes reasoning about such signals to be within the capacities of LLMs, even if they fail at the task at hand. 

\autoref{fig:respect} illustrates the \respect process: the model interacts with humans to accomplish tasks, retrospectively examines its own past interactions, and then re-trains. 
This process progresses in rounds, alternating between interaction and training, improving model task capability over time.
Critically, unlike common recipes for training from human feedback, \respect does not require any external annotation~\cite[RLHF]{ouyang2022Training} or even soliciting feedback from the users themselves~\citep{suhr2023Continual}.

We deploy \respect in \multiref over multiple rounds of grounded interactions with human users and re-training. We use \Idefics \citep{laurencon2024What} as our MLLM, and experiment with multiple learning algorithms, including filtered fine-tuning (FFT), REINFORCE-style policy gradient~\citep{williams1992Simplea, kojima2021Continual}, and KTO~\citep{ethayarajh2024KTO}. 
Across our experiments, we observe that \Idefics  effectively decodes feedback, even as it initially performs poorly at the task. 
In our longest running experiment, we observe model task completion rate improves from 31\% to 82\%. 
Our code, data, and models are available at \href{https://lil-lab.github.io/respect}{\texttt{https://lil-lab.github.io/respect}}.

\section{Technical Overview and Notation}

We conduct our study in an interaction scenario called \multiref (\autoref{sec:multiref}). Each \emph{interaction} is a \emph{game} that involves multiple turns. At each \emph{turn}, the speaker first produces a free-form natural language text, and then 
the listener performs an \emph{action} according to a \emph{policy}. 
Zooming out of the mechanics of the interaction, we experiment with a continual learning setup (\autoref{sec:respect}).
Our study progresses in \emph{rounds}. In each round, the MLLM is first deployed to interact with users and complete tasks, and then the interactions are used to re-train the  policy's model.  Multiple rounds enable us to observe the long-term dynamics of learning from the model's own interactions. This includes the robustness of our feedback decoding and training methods to the changing distribution of the data likely to be seen in an adaptive system in the wild. 

\paragraph{Task Notation} 

The MLLM policy's task is to respond effectively to human utterances given in conversational context. 
Formally, let $\policy_\param(\action_t \vert \context_t)$ be the $\param$-parameterized MLLM policy, with $t$ being the current interaction turn, $\action_t$ an action string that represents the model response, and $\context_t$ being the context on which the policy is conditioned. 
The context includes the interaction history up to and excluding turn $t$, including current (i.e., at turn $t-1$) and past user utterances, as well as any other relevant context in which the interaction takes place.

\newcommand{\prob}{p}

\paragraph{Learning and Deployment}
Each round $\round$ includes a deployment, followed by training on the interactions between the deployed model and humans (\autoref{fig:respect}).
During deployment at round $\round$, the model $\policy_{\param_\round}$ interacts with users. 
For each model action $\hat{\action}_t \sim \policy_{\param_\round}(\action \vert\context_t)$, we record a tuple $(\context_t, \hat{\action}_t, \prob_t, \futureutt_t)$, where $\context_t$ is the context given to the model at time $t$ to predict action $\hat{\action}_t$, $\prob_t$ is the probability of $\hat{\action}_t$ at the time of prediction, and $\futureutt_t$ is the remainder of the interaction following $\hat{\action}_t$. 
Critically, we do not solicit human feedback during the interaction or after it. 
We compute the implicit feedback $\hat{\feedback}_t$ using a feedback decoder $\feedbackdecoder(\context_t, \hat{\action}_t, \futureutt_t)$, to obtain tuples $(\context_t, \hat{\action}_t, \hat{\feedback}_t, \prob_t)$. 
We experiment with three learning objectives using this feedback: filtered fine-tuning (FFT), policy gradient, and KTO.

\paragraph{Evaluation} We measure the quality of the model $\policy_{\param_\round}(\action_t|\context_t)$ at each round $\round$ primarily by interaction success rates from live human-model deployments. The same interactions are used to train the model for the next round. 
We track other metrics, such as the number of turns per interaction as an efficiency measure.
We also annotate a subset of the interactions post hoc to measure utterance-level policy success rate and feedback decoder accuracy.

\section{\multiref: a Multi-turn Grounded Interaction Scenario}\label{sec:multiref}

\begin{figure}
\centering
\includegraphics[trim={15 350 1100 37}, clip, width=0.95\linewidth]{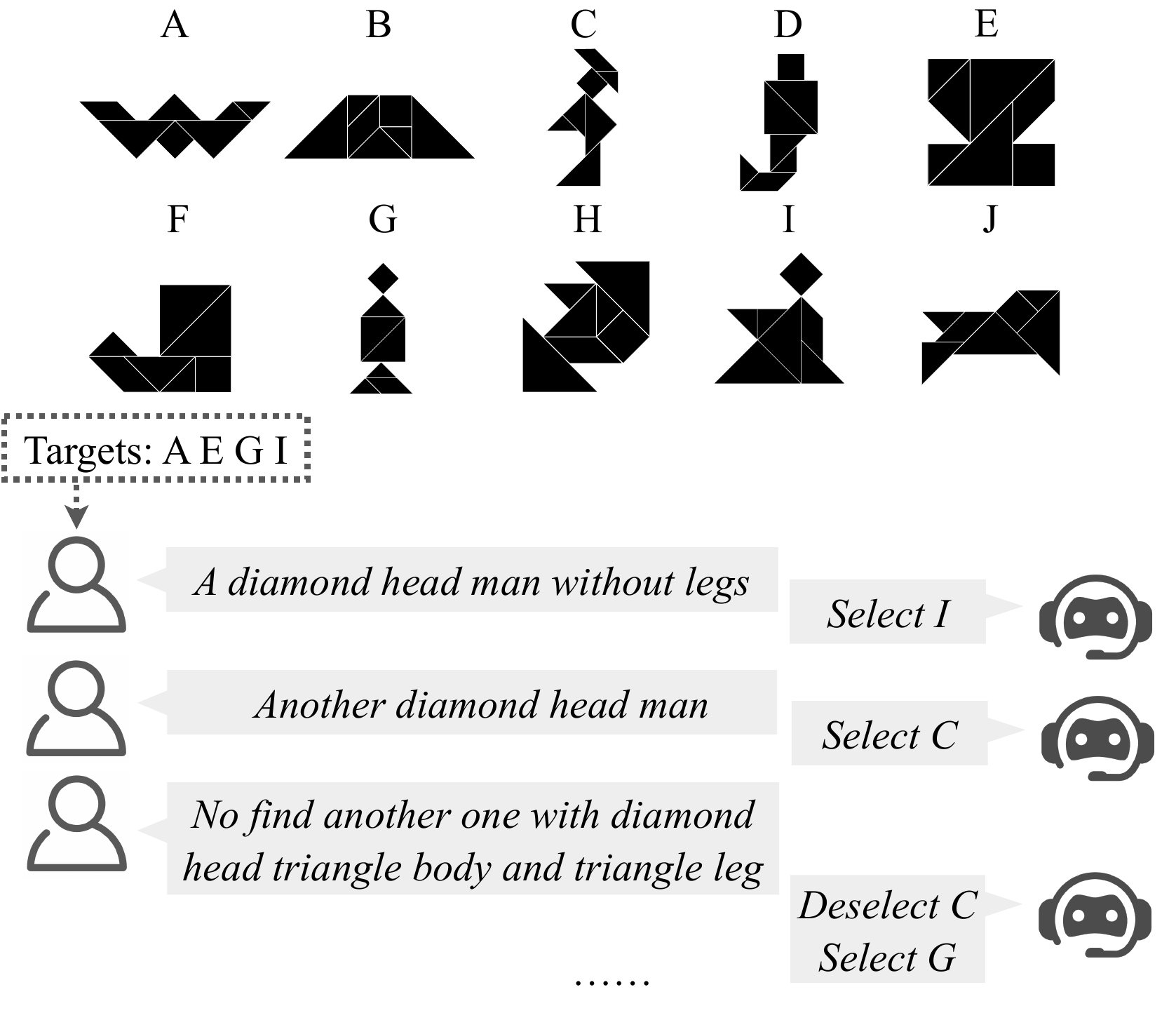}
\vspace{-5pt}
\caption{The interaction scenario we use in our experiments. 
\multiref is a multi-turn reference game. A speaker and a listener both observe a shared set of tangram shapes in different orders. The speaker describes targets for the listener to select, often gradually over multiple turns.
As an interaction progresses, the speaker naturally produces implicit feedback signals that validate or reject the listener's actions.}\label{fig:interaction}
\vspace{-5pt}
\end{figure}

Key to our study is that tasks are relayed gradually across multiple turns, as often occurs in human interaction. 
We create \multiref, a conversational interaction scenario where two partners, a \emph{speaker} and a \emph{listener}, coordinate on the selection of a set of items (\autoref{fig:interaction}). 
In our studies, the speaker is always a human, and the listener is a model.
\multiref generalizes the commonly studied reference game scenario. Its design and our choice of stimuli are grounded in existing work from both cognitive science and computational language modeling~\citep{rosenberg1964Speakers, clark1986Referring, schober1989Understanding, goodman2016pragmatic}.
Both partners observe a shared set of images, but in different order. 
The speaker is given a subset of the images as targets, with the goal of communicating the targets to the listener, so the latter selects the exact subset. 
Only the speaker can write natural language messages and only the listener can select or deselect images, by generating a structured string (e.g., \code{Deselect E select F}).
The interaction concludes successfully once all and only targets are selected, or fails on timeout, 20 turns in our studies.

\multiref is both accessible to crowdsourcing workers and encourages constructing the solution in multiple turns, thereby creating multi-turn interactions that likely include the learning signals we aim to study. 
The rules are simple: the speaker describes targets for the listener to select. 
This makes \multiref easily accessible to crowdsourcing workers.
At the same time, the solution the speaker communicates is relatively complex, because of the large solution space.  
In conventional reference games, the listener's goal is to select one image from $n$ images, so the number of possible solutions is $n$. 
In \multiref, the goal is to select a subset of unknown size from $n$ images, so the combinatorial solution space is exponential in $n$. 
Meanwhile, the solution is decomposable, 
creating natural opportunities for gradual instruction and implicit, immediate, and incremental feedback.

We use tangram shapes from the diverse \kilogram dataset~\citep{ji2022Abstract}. Tangrams are abstract shapes that are designed to elicit common concepts in humans. 
This abstractness often leads to ambiguous descriptions open to interpretation, e.g., Shape~A in \autoref{fig:interaction} can be described as a \nlstring{bat}, a \nlstring{lowercase w}, or even a \nlstring{star wars star fighter}.
Tangrams naturally provide an ambiguous and challenging stimuli for human interaction~\citep{clark1986Referring, schober1989Understanding,foxtree1999Listening,hawkins2020Characterizing}, thereby leading to highly diverse language. They also remain challenging for contemporary MLLMs to reason about~\citep{ji2022Abstract,cheng2024comt}, leaving significant room for learning.

The free-form natural language human speakers produce in \multiref is very diverse, and balances between competing pressures. 
First, it often requires complex pragmatic reasoning~\citep{clark1986Referring,schober1989Understanding,horton2002Speakers}, because of the abstractness of tangrams. 
This is compounded by how the combinatorial solution space drives humans to balance between relaying as much information as possible, and relaying clear objectives to make gradual progress -- a balance between two Gricean maxims: quantity and manner \citep{grice1975logic}. 
Speakers may include explicit feedback such as \nlstring{good}, or \nlstring{deselect the last one}; the speaker may describe two targets in a single utterance (\nlstring{select two men}); speakers may refer to previous selections without directly describing targets (\nlstring{the other one} or \nlstring{try again}).
\aautoref{case-studies} shows several interaction case studies.
Diverse language with abstract stimuli poses a challenging reasoning problem for the listener model.

\multiref is not designed to arbitrarily increase complexity, but to naturally expose core aspects of human communication.
Yet the scenario is controlled and scoped, allowing for easy measurement of task completion and progress, and making learning feasible with relatively limited crowdsourcing. 
This makes \multiref particularly suitable for research in academia or other low-resource settings.

\section{\respect: Retrospective Learning from Past Interactions}\label{sec:respect}

\respect has two components: decoding implicit feedback from past interactions (\emph{retrospection}) and learning from the decoded feedback signals (\emph{learning}).
We deploy \respect in an iterative continual learning setup to observe the dynamics of learning from the models' own interaction over time.
However, the method itself is not limited to continual learning, and can be applied a single step.

Formally, \respect re-estimates the policy parameters $\param$ given on-policy interactions.
We assume access to a raw dataset $\dataset^{\text{raw}} = \{(\context^{(i)}, \hat\action^{(i)}, \prob^{(i)}, \futureutt^{(i)})\}_{i=1}^N$, where $\context^{(i)}$ is the policy context consisting of images and the conversation so far, $\hat\action^{(i)}$ is the predicted action, $\prob^{(i)}$ is the probability of this action, and $\futureutt^{(i)}$ is the remainder of the interaction following $\hat\action^{(i)}$.\footnote{For simplicity, we omit the turn index in this section.} A single interaction is split into multiple training datapoints.\footnote{Multiple selections or deselections in one turn such as \code{Deselect E select F} are considered a single action.}
In our continual learning setup, $\dataset^{\textrm{raw}}$ is a union of all data collected from past rounds. 
The feedback decoder $\feedbackdecoder$ computes a categorical feedback $\hat\feedbacktype^{(i)} \in \{\text{\feedbackpos, \feedbackneu, \feedbackneg}\}$ for each action $\hat\action^{(i)}$ holistically based on its context $\context^{(i)}$, action taken $\hat\action^{(i)}$, and follow-up utterances $\futureutt^{(i)}$.\footnote{We do not compute feedback for the last action in each interaction because there is no follow-up interaction. They are excluded from $\dataset^{\textrm{raw}}$.}
This process transforms $\dataset^{\textrm{raw}}$ to $\dataset = \{(\context^{(i)}, \hat\action^{(i)}, \prob^{(i)}, \hat\feedback^{(i)})\}_{i=1}^N$. 

\subsection{Decoding Implicit Feedback} %

We implement the feedback decoder $\feedbackdecoder$ by prompting the model to analyze past interaction tuples $(\context, \hat\action, \prob, \futureutt)$ to compute feedback $\hat\feedback = \feedbackdecoder(\context, \hat\action, \futureutt)$ (\autoref{fig:feedback-decoder-prompt}).
We hypothesize that pretrained LLMs have the ability to reason about the relatively constrained space of implicit feedback signals, even if they fail at the task.
Critically, this process does not rely on a stronger LLM for critique nor on past interactions created by other LLMs, ruling out concerns about distillation.
We experiment with binary or ternary feedback. 
The feedback decoder is designed to identify general linguistic cues, and not for the specific task we study. 
We assume no access to any auxiliary annotation or privileged information (e.g., no access to selection ground-truth or interaction success). 
This assumption of no auxiliary labels or human annotation enables us to explore the most general scenario of learning from interactions, and our method scales freely as deployment data streams in.

\newcommand\bcolor{c1}
\newcommand\tcolor{c2}

\begin{figure}[t]
    \footnotesize
    \centering
    \begin{tcolorbox}[width=0.48\textwidth, boxsep=0pt, left=4pt, right=4pt, top=4pt, bottom=4pt, sharp corners]
        User: Please carefully read the following conversation and answer: Is the very last utterance from the speaker \textcolor{\bcolor}{positive or negative} \textcolor{\tcolor}{positive, neutral, or negative} feedback? Often negative feedback include corrections and keywords like no, not, undo, don't, with generally negative sentiment, while positive feedback often includes good, yes, correct, okay, or simply move on to the next stage. \textcolor{\bcolor}{Lean towards negative if it sounds neutral.}\\
        (start of the conversation)

        Listener: Deselect F select G
        
        Speaker: yes, pick the thin person with a triangle head
        
        Listener: Select A \boxcomment{Action to focus on}
        
        Speaker: yes, pick the house with chimney  \boxcomment{Feedback}

        (end of the conversation)\\
        Answer a single word, \textcolor{\bcolor}{Positive, or Negative} \textcolor{\tcolor}{Positive, Neutral or Negative}.\\
        Assistant: \textbf{Positive}
    \end{tcolorbox}
    \vspace{-10pt}
    \caption{The text-only prompt used to decode feedback from past interactions. This figure combines the prompts for both binary and ternary feedback decoding. \textcolor{\bcolor}{Green}: binary case only. \textcolor{\tcolor}{Orange}: ternary case only. The verbal \textbf{feedback generated by the model} is in bold. Additional \emph{\textcolor{c0}{comments for readability}} are in blue italics.}\label{fig:feedback-decoder-prompt}
    \vspace{-5pt}
\end{figure}

\subsection{Learning}

We study three approaches to learn from the processed dataset $\dataset = \{(\context^{(i)}, \hat\action^{(i)}, \prob^{(i)}, \hat\feedbacktype^{(i)})\}_{i=1}^N$.

\paragraph{Filtered fine-tuning}

We fine-tune on positive data points ($\hat\feedback^{(i)} = \feedbackpos$) and reject data points with decoded $\feedbackneu$ or $\feedbackneg$ feedback. We use cross entropy loss, with label smoothing to regularize 
 learning. 
This method follows approaches that filter to retain examples to reinforce, but relies on decoded feedback, rather than, for example, verifiers~\cite{zelikman2024quiet}.

\paragraph{Reinforcement Learning}
We use simple REINFORCE-style policy gradient~\citep{williams1992Simplea}. 
This choice has been shown as effective in few-sample regimes, as in studies with human users~\cite{kojima2021Continual,suhr2023Continual}. 
It avoids data demanding value function estimation and plurality of hyperparameters, both downsides of algorithms like PPO~\cite{schulman2017Proximal}.  
Recent work~\citep{ahmadian2024Back} also suggests REINFORCE can produce on-par results in LLMs with PPO despite its simplicity.
Because of the few-sample regime, we do not have sufficient data to estimate a reward function, so cannot perform online RL~\cite{ouyang2022Training}. 
We train in an offline fashion within each individual round. The iterative rounds of training and deployment create a hybrid offline-online process.

We map the categorical feedback generated by the decoder $\feedbacktype^{(i)}$ to numerical rewards:
\begin{equation}
    R(\feedback) = 
    \begin{cases}
      1 , & \feedback = \feedbackpos \\
      0 , & \feedback = \feedbackneu \\
      -0.1 , & \feedback = \feedbackneg
    \end{cases}\;\;.
\end{equation}
Dropping the $i$-superscripts for simplicity, the gradient estimator for a single example is:
\begin{equation}
\begin{aligned}
    \frac{\partial}{\partial\param} &= c R(\hat\feedback) \nabla \text{log}P(\hat\action | \context; \param) \\
    c &= 
    \begin{cases}
      1 , & \text{if }R(\hat\feedback) \geq 0 \\
      \frac{P(\hat\action | \context; \param)}{\prob}, & \text{if } R(\hat\feedback) < 0 \;\;,
    \end{cases}
\end{aligned}
\end{equation}
where the coefficient $c$ downweights examples with negative reward by their inverse propensity score~\citep{kojima2021Continual}. This is critical because $\lim_{P(\cdot) \rightarrow 0} \text{log}P(\cdot) = -\infty$.

\paragraph{Utility Maximization} We use Kahneman-Tversky Optimization~\citep[KTO;][]{ethayarajh2024KTO}. 
KTO fits our scenario because it assumes per-example binary human feedback, in contrast to methods like DPO that require pair-wise preferences~\cite{rafailov2023direct}. 
We consider examples with decoded $\feedbackpos$ feedback as \emph{desired} utterances, those with decoded $\feedbackneg$  feedback as \emph{undesired}, and discard those with $\feedbackneu$ feedback.
We refer readers to \citet{ethayarajh2024KTO} for the definition of the objective.

\section{Experimental setup}\label{sec:experiment}

\paragraph{Interaction Instantiation}

We use the \kilogram~\citep{ji2022Abstract} tangram images~\cite{gul2024CoGen}.
\kilogram contains 1{,}013 images. We randomly split them into a main split (912 tangrams) and a development split (101 tangrams). 
We create interaction contexts by randomly sampling 10 tangrams,  and randomly selecting 3--5 as targets (\aautoref{game-design}). 
The development split is exclusively used for seeding the initial listener policy $\policy_{\param_0}$, and all human-bot interactions are conducted on images from the main split, i.e., tangrams that the seed policy $\policy_{\param_0}$ has never seen before.

\paragraph{Model and Initialization} 
We use \Idefics \citep{laurencon2024What} for both the policy and feedback decoder. 
We fine-tune with LoRA~\citep{hu2021LoRAa}.
We seed the initial policy $\policy_{\param_0}$ by fine-tuning the pretrained \Idefics weights on a small dataset from 25 human-human games constructed with the development split tangrams. \aautoref{learning-details-seeding} provides further details. 
$\dataset_0$ is reused in continual training via rehearsal.
We use the original \Idefics for feedback decoding, because our limited data is likely to inhibit general linguistic knowledge. This means we cannot benefit from improvements in the model feedback decoding, likely low-balling the potential of the approach.\footnote{It remains an important direction for future work to keep the decoder model in sync with the policy.}
The original \Idefics provides robust feedback decoding out of the box, confirming our hypothesis, and providing a solid ground for our experiments. 

\paragraph{System variants}
We study six system variants along two dimensions: feedback decoder configuration (binary \textsc{b} vs. ternary \textsc{t}) and optimization methods (FFT vs. RL vs. KTO): 
\bp and \tp train on positive data points with an FFT objective (\textsc{fft}).
\ba and \ta trains on both positive and negative data points using RL.
\bk and \tk are like \ba and \ta, but using KTO.

For variants involving negative data points (\ba, \ta, \bk, and \tk), we subsample negative ones to keep the positive:negative ratio close to 5:4~ \citep{ethayarajh2024KTO}.

\paragraph{Deployment} We conduct three rounds of training-deployment for all six systems and three more rounds for the top system, \bp, to observe its progress over a longer period.
This cascaded design is a direct consequence of the high cost of crowdsourcing.
We do not distinguish between training and evaluation in the traditional sense. Instead, all listener policies are evaluated live on MTurk on about 330 human-bot interactions each round containing roughly 2{,}400 turns. The same data is used to train the next iteration of policies. The policies in the same round are deployed concurrently in a randomized experiment on the same set of games to mitigate bias due to game difficulty. 
\aautoref{mturk-details} provides further details.

\paragraph{Learning Implementation Details}

We use the validation set for model selection throughout continual learning.
Following prior work~\citep{misra2017Mapping, muller2019When,liu2022FewShot}, we add an entropy term and length normalization to all three objectives to reduce over-fitting given the relatively small amount of data.
\aautoref{learning-details} provides additional reproducibility details.
Unlike FFT and RL, where we train from scratch each round, when using KTO, we continually fine-tune from a previous model checkpoint $\param_{\round}$ to obtain $\param_{\round+1}$ with data accumulation. This was shown to outperform training from scratch in pilot studies.

\paragraph{Evaluation} We evaluate each system variant at each round by the success rate during the live deployment. 
We report both interaction- and utterance-level success rates. 
The interaction level success rate is straightforward - whether the game ended with all targets selected by the listener and nothing else. 
We do not have access to utterance-level ground truth (i.e., the intended action) to compute success rate, so we sample 1{,}000 utterances per round from \bp to annotate by MTurk workers post-hoc. 
We report two measures: exact match between the annotation and model action and similarity score, which is based on the computed similarity between the tangrams selected or deselected during the turn by the human annotator and the system.
We evaluate the feedback decoder by comparing its predictions with human interpretations collected during the post-hoc annotation. 
Lastly, we track the number of turns per interaction. 
\aautoref{learning-details-metrics} provides metric definitions. 

\section{Results and analysis} \label{sec:results}

We deploy our models for three rounds, with additional three rounds for \bp, the best-performing variant, to better understand long-term dynamics. 
All our results are from concurrent randomized deployments, where the models interact with humans in real time. 
We collect a total of 7{,}230 interactions consisting of 55{,}004 utterances over four weeks, at a cost of \$11{,}180 USD.
\footnote{This crowdsourcing cost is for conducting the controlled experiment. There are no data costs when applying \respect on a deployed system, because learning signals arise from interactions, not from external annotations.}

\begin{figure}
    \centering
    \includegraphics[width=0.49\textwidth, trim={0 0.5in 0 0}, clip,]{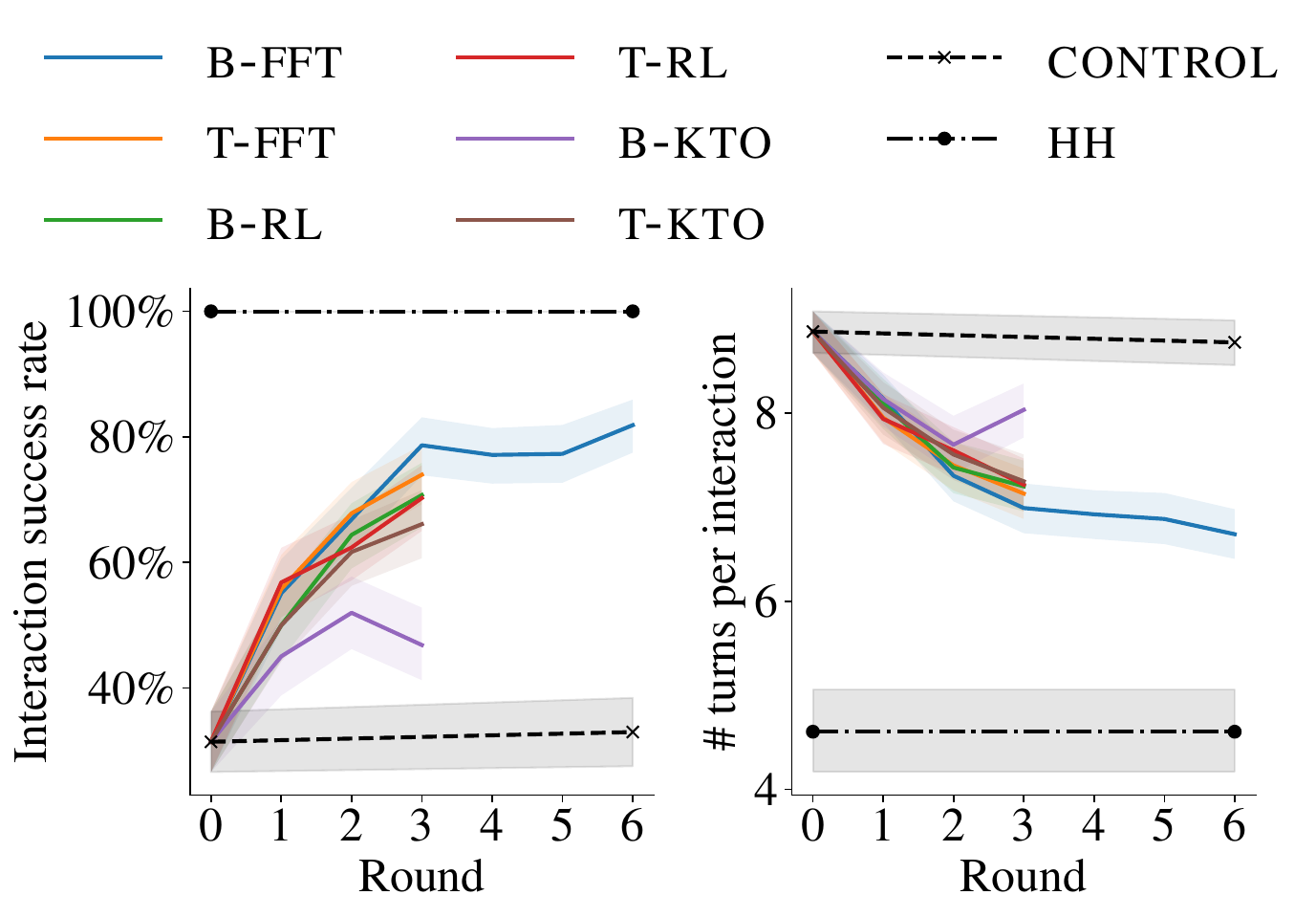}
    \vspace{-10pt}
    \caption{Task performance and efficiency improve as the policy learns from more past interactions. We present deployment results across three rounds for six concurrent systems, and three more rounds for the best system \textcolor{c0}{\bp}, together with human-human references (\textsc{hh}) and a redeployment of the initial policy $\policy_{\param_0}$ (\control). \emph{Left:} interaction-level success rate ($\uparrow$, higher is better). \emph{Right:} interaction-level efficiency by \# turns per interactions ($\downarrow$). Shades are 95\% confidence intervals by bootstrapping with 10{,}000 resamples.}\label{fig:task-performance}
    \vspace{-5pt}
\end{figure}

\begin{figure*}[t!]
    \vspace{5pt}
    \centering
    \includegraphics[width=0.8\linewidth]{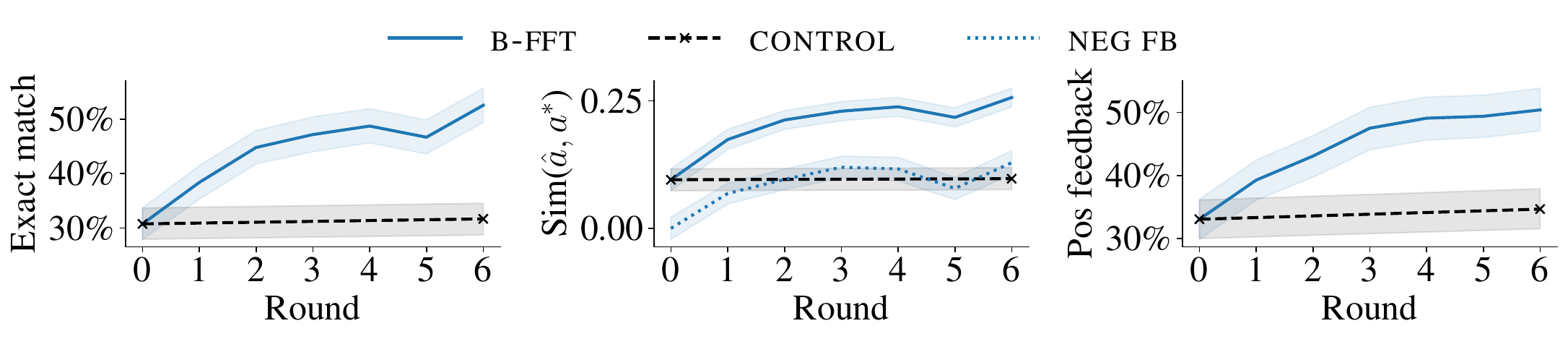}
    \vspace{-5pt}
    \caption{ 
        Turn-level performance of \bp evaluated by post-hoc human annotations. \emph{Left:} \% turns where the policy's action $\hat \action$ matches exactly the human listener's action $\action^*$ ($\uparrow$). 
        \emph{Center:} similarity between the policy's action and the human listener's action ($\uparrow$). Even actions that receive negative feedback in deployment (\textsc{neg fb}) are increasingly similar to human actions.
        \emph{Right:} \% policy actions annotated to have received positive implicit feedback from human listeners ($\uparrow$). Shades are 95\% confidence intervals by bootstrapping with 10{,}000 resamples.
    }\label{fig:human-annotations}
    \vspace{-5pt}
\end{figure*}

\autoref{fig:task-performance} shows deployment statistics for all six system variants, as well as control deployments for the initial policy and human-human games.\footnote{We present results in rounds for simplicity. \aautoref{cumsum} connects rounds to cumulative number of interactions. \aautoref{tab-results} presents full tables corresponding to these plots.}
\autoref{fig:human-annotations} shows utterance-level statistics for \bp from the post-hoc annotations we collected for evaluation.
The interaction success rate of all systems improves monotonically in the first three rounds, except for \bk in round 3. \bp plateaus in rounds 4 and 5, before resuming its improvement. \footnote{The reasons behind the plateau are hard to infer. One hypothesis is that changes in the amount of data over time made some settings sub-optimal. We conducted a separate deployment, branching out from round 3 with \bp and more expressive LoRA adapters. This increase in expressivity allows the model to continue its monotonous improvement (\aautoref{lora-enhanced}). This mini experiment illustrates the complexities of continual learning with current learning systems.}
Figures \ref{fig:case-study-triplet-14-initial}--\ref{fig:case-study-triplet-14-human-human} in the appendix show example final-round interactions comparing the initial and final models, as well as human listeners.

Overall, \bp improves interaction-level success rate by 51\% (31\%$\rightarrow$82\%) and utterance-level exact match by 22\% (31\%$\rightarrow$53\%).
At the last round, following the plateau, \bp interaction success rate improves by 5\% (77\%$\rightarrow$82\%).
The number of turns follows these trends. As the policy gets better, more games are completed within the allotted number of turns, and even faster. 
\bp starts with 8.9 turns per game, and concludes with 6.7 per game.
The center panel of \autoref{fig:human-annotations} shows that actions taken by the policy increasingly resemble human actions, even mistakes (actions that receive negative feedback) become more similar to human actions. 
All other statistics largely track these, except some of the utterance-level statistics around when \bp plateaus. 
While all show a deviation from the monotonous earlier trend, some show a temporary decrease and not just a stagnation, but delayed by one round. 
This illustrates the complex dynamics of continual learning.

There remains a gap between \bp (our leading system) and \hh (human-human interactions) even with the same worker pool. \hh interactions show perfect task success rate and high efficiency. Our intuition is that the gap is due to the lack of long-term credit assignment in our learning method. This is especially influential in learning to reason about later turns, where credit assignment to the longer history is more complex. 
Excluding some past turns (i.e., sliding window approach) may address this issue.
This learning challenge is compounded by data scarcity: we have significantly less data for later turns.

\paragraph{User adaptation}

A potential confounding factor to the improvement in interaction success rate is users adapting to the interaction scenario and the model, instead of policy improvement~\citep{hawkins2020Continual}. 
We redeploy the initial policy $\policy_{\param_0}$ concurrently in the final \bp round to test this (\control in \autoref{fig:task-performance}). 
The interaction success rate of \control remains unchanged over time (31\% $\rightarrow$ 33\%), suggesting speaker adaptation do not explain the overall 51\% absolute improvement in \bp task success rate.

\paragraph{Positive Only vs. All data}

The difference between systems using positive learning signals only (\bp, \tp) and those using all (\ba, \ta, \bk, \tk) is in learning objectives (FFT/RL/KTO). 
Overall, the systems based on positive signals only perform better. 
It is expected that positive signals will be more informative for learning. Our policy acts in a large action space. Negative rewards suppress specific actions, but without more information about what a good action is, they simply encourage a uniform distribution. 
This has been shown to have a helpful regularizing effect in ~\citet{kojima2021Continual}. 
However, not only does negative feedback not help meaningfully, it seems to confuse the learner. The positive-only systems that, in effect, have access to fewer learning signals perform better. 
Utilizing negative signals better is an important direction for future work.

\paragraph{Feedback Decoder Quality}

\begin{figure}[t]
    \centering
    \includegraphics[width=0.47\textwidth, trim={0 0.1in 9.5in 0}, clip]{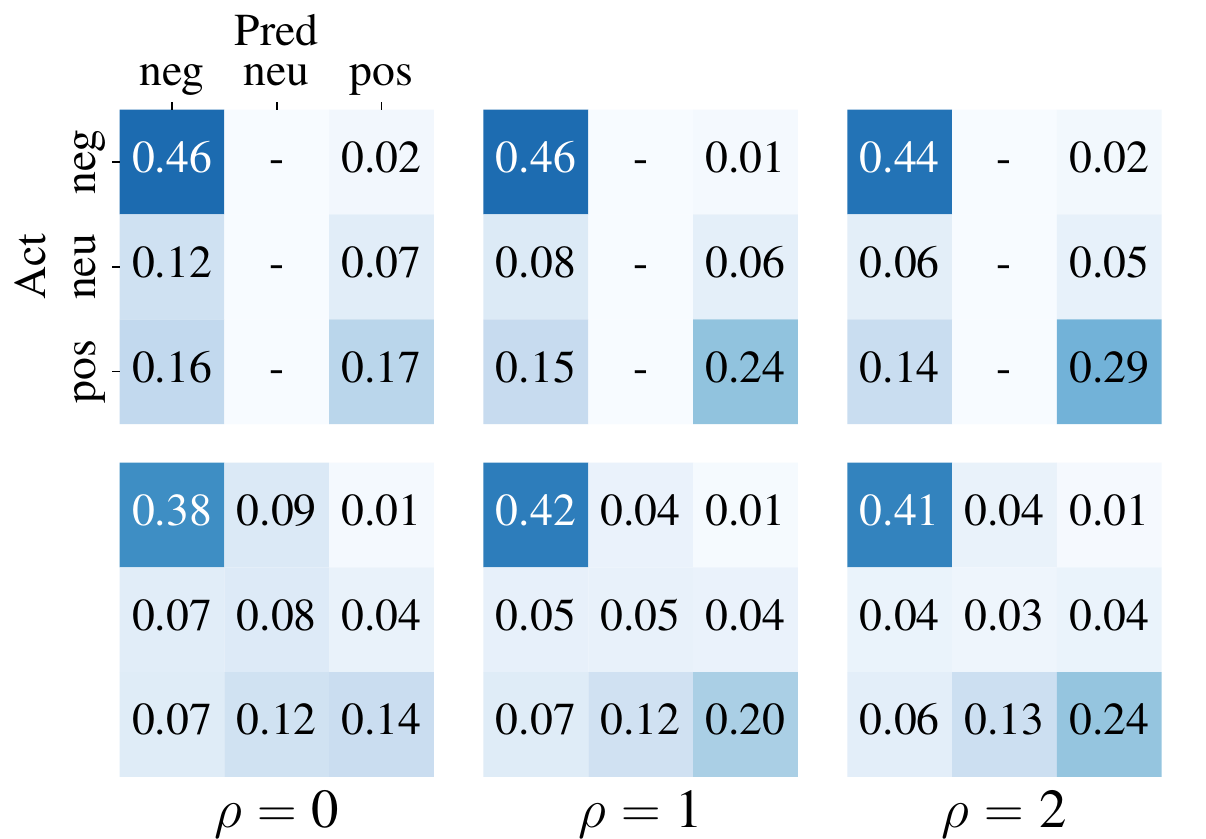}
    \vspace{-5pt}
    \caption{Confusion matrices of the binary (top) and ternary (bottom) feedback decoders over rounds. Feedback decoders yield negligibly low false positives (top right corner). The feedback decoder also correctly classifies more than 60\% (diagonals) across rounds.}\label{fig:rd-confusion-matrix}
    \vspace{-5pt}
\end{figure}

We evaluate the feedback decoder through our post-hoc annotation task.
Workers annotate each turn if the speaker was satisfied with the answer given their follow-up utterances. 
The feedback decoder performance is stable throughout the rounds, showing robustness to changes in the data distribution (\autoref{fig:rd-confusion-matrix}).
We observe above 90\% precision consistently, after combining actual positives and neutrals.
The ternary feedback decoder is more conservative and labels more positive turns as neutrals. This is a task-dependent trade-off. The zero feedback of neutrals essentially eliminates the examples, but allows for slightly cleaner data. We empirically observed it is beneficial to have slightly noisy data but more of it.

\paragraph{FFT vs. REINFORCE vs. KTO}

Overall, the FFT (\bp and \tp) perform best. 
The KTO variants (\bk and \tk) trail after the REINFORCE variants (\ba, \ta). \bk even diverges at some point. 
We suspect that the KTO recipe struggles in the continual optimization scenario, where the model is fine-tuned multiple times. 
We observe that \bk deteriorates in rounds 2 and 3, and starts generating illegal outputs (e.g., \code{Deselect select}). \aautoref{learning-details-hyperparameters} describes a quick intervention we applied to mitigate this issue. Although it eliminated the illegal outputs, the quality remained low. 
Further refinement of KTO or hyperparameters may help, however, this is a complex process in a live deployment.

\begin{figure}[t]
    \centering
    \includegraphics[width=0.47\textwidth]{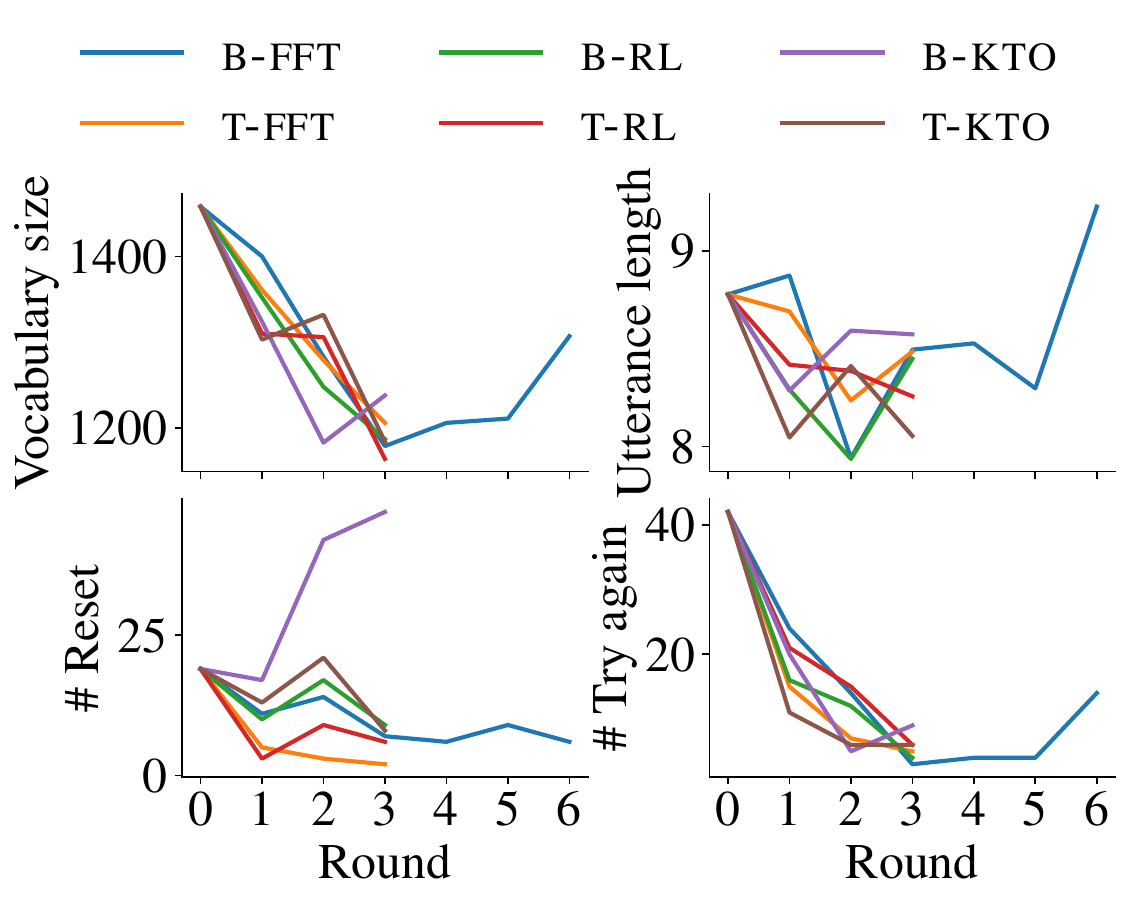}
    \vspace{-10pt}
    \caption{Language analysis of human instructions. All systems see a decrease in instruction complexity in the first three rounds, except for \bk, suggesting adaptation on the speaker's side, as expected when humans become more familiar with the domain. Reset/frustration signals drop, a reflection of the model improving.}\label{fig:language-analysis}
    \vspace{-10pt}
\end{figure}

\paragraph{Language analysis}

Human instructions changes over time (\autoref{fig:language-analysis}). 
We observe a reduction in vocabulary size and utterance length early on. This is expected, and follows known observations in how humans adapt to reduce cognitive costs~\cite[e.g.,][]{clark1986Referring,effenberger2021Analysisb}. 
However, in later rounds, \bp witnesses an increase in vocabulary size and utterance length. 
This surprising trend reversal is attributed to three outlier workers, so does not reflect a change in population behavior. 
Reset signals drop, an indication of improved collaborated task performance. These trends are fairly consistent across system variants, except for \bk, which also diverges in performance. 
Initially workers tend to use \nlstring{Try again} instead of directly describing a target, or request a reset with instructions like \nlstring{Deselect everything}~(\autoref{fig:case-study-try-again} and \autoref{fig:case-study-reset}). The occurrences of both decrease in later rounds. 
Even though workers change their language, this does not influence the initial policy's $\policy_{\param_0}$  performance (\autoref{fig:task-performance}).

\section{Related work}\label{sec:related}

\paragraph{Learning from feedback}

RL from human feedback~\citep[RLHF;][]{ouyang2022Training}, the most common recipe to learn from feedback, relies on soliciting pair-wise preferences from annotators, while we rely on \emph{unpaired} signals from the interaction itself.
Learning from explicit feedback on a single output has also been studied, either in the form of binary feedback~\citep{ethayarajh2024KTO,suhr2023Continual, gao2023Continually} or through more expressive editing~\citep{gao2024Aligning} or commenting and refinement~\citep{li2017Dialogue, sumers2021Learning, scheurer2023Training}.
\citet{hancock2019Learning} pauses interactions by consulting a separately trained satisfaction predictor, and solicits explicit feedback.
We do not solicit explicit feedback, but rely on natural signals that arise from the follow-up interaction.

\paragraph{Learning from naturally occurring signals}

\citet{kojima2021Continual} learns to generate instructions by observing how humans follow them. 
\citet{pang2023Leveraging} maximizes heuristics, such as longer responses from humans. 
\citet{Artzi:11} studied the use of naturally occurring recovery efforts. 
Instead, we opt for a general approach to infer feedback from natural interactions.
Concurrent work by \citet{don-yehiya2024Learning} uses similar linguistic cues to ours, basing their reward decoder on the taxonomy of \citet{petrak2023Learning}. 
They focus on a distillation-like scenario, where the interactions they learn from originate from other models, many of which are stronger than the model they train. 
We focus on self-improvement, where it is critical that no stronger model is involved. 
Our works complement each other and strengthen our conclusions. Their work shows our signal of interest can be derived from large-scale diverse data, whereas we show a single-model loop can use this signal to drive improvement over time.

\paragraph{LLMs that self-\relax improve} 
A common approach to improve models is via AI feedback, solicited from the model itself or another model~\citep{bai2022Constitutional, burns2023WeakStrong, madaan2023SelfRefine, kumar2024training, qu2024Recursive, yuan2024SelfRewarding, li2024Selective}, or via a hand-crafted verifier~\citep{anil2021Learning, kirchner2024ProverVerifier}. 
In contrast, we rely on implicit \emph{real human} feedback from deployment interactions. 
This signal is less influenced by model biases or limitations, and does not require a model to validate the task. 
Our approach can be combined with any of the above, and allows to leverage deployment interactions for free.

\section{Conclusion}

We study the ability of models to decode implicit feedback from  interactions with humans, and the efficacy of this learning signal. 
We operationalize learning from this signal as retrospective learning, an annotation-free approach that leverages signals from naturally occurring feedback in interactions.
We demonstrate its effectiveness in long-term human-in-the-loop deployments and robustness to variants. We hope to unveil the potential of a common yet underutilized learning signal and eventually inspire an evolving language model that learns continuously without any expert annotation.

\section*{Limitations}
We design \multiref to study real interactions over a period of time, as opposed to evaluating on a static benchmark. We make trade-offs between the generality of the task, its fit to our experimental questions, and the ability to iterate on a prototype fast, and without high costs.
\multiref exposes the problem we aim to study and provides an experimental ground to show a strong, measurable effect, without requiring prohibitive amount of data. 
Any such data requirement would make our deployment impossible in research settings, thereby not allowing us to observe the dynamics of interacting with real humans over a period of time. 
That said, the cost of this choice is that our models specialize to the task of \multiref. They do not improve their abilities in ways that generalize beyond \multiref, and likely even experience an erosion of these capabilities. 
While our setup allows us to provide a clean and significant answer to our research questions, it is important to expand our study to other domains, such as summarization or conversational question answering, where similar signals may be more complex, farther apart, or demand long-term credit assignment. 
Our method uses only a scalar reward. Another interesting orthogonal direction is expanding the expressivity of the feedback decoder, such that it recovers a more expressive signal (e.g., a natural language explanation). 

\section*{Ethical considerations}
The ultimate application of our method is an automated pipeline that continuously learns from human-model interactions in deployment without additional human annotations. 
A naive implementation of our method, like any other approach that learns from human feedback data in deployment, is subject to data poisoning and potentially learning harmful behaviors when, for example, a malicious user verbally incentivizes harmful answers. 
In general, data quality assurance is essential to safeguard learning from human feedback in deployment.

\section*{Acknowledgments}

This research was supported by NSF under grants No. 1750499 and OAC-2311521, NASA under award No. 20-OSTFL20-0053, a gift from Open Philanthropy, a gift from Apple, the National Artificial Intelligence Research Resource (NAIRR) Pilot, the Frontera supercomputer supported by the National Science Foundation (award NSF-OAC 1818253) at the Texas Advanced Computing Center (TACC) at The University of Texas at Austin, and the Delta advanced computing and data resource which is supported by the National Science Foundation (award NSF-OAC 2005572). 
GG was supported by an NSF REU supplement for NSF  grant No. 175049. YC was supported by Bowers Undergraduate Research Experience program. 
Any opinions, findings and conclusions or recommendations expressed in this material are those of the author(s) and do not necessarily reflect the views of the National Science Foundation, NASA, or the other funders.

\paragraph{Author Contributions}
ZC developed and implemented the methods, trained models, conducted pilot studies, and the continual learning study. MOG contributed code for tangram reference games and designed the human evaluation study. 
MOG, AW, YC, and GG designed the \multiref scenario. 
YC and GG implemented the \multiref interface, recruited MTurk workers, and collected human-human interactions, with MOG and AW's mentorship. ZC, MOG, and AW contributed to the writing. YA advised and oversaw all parts of the project.

\bibliography{multiref}

\clearpage
\appendix
\section{The \multiref Game Design and Data Collection}\label{game-design}

\subsection{Interaction Design}

\multiref is a multi-target, multi-turn reference game between two players, a \emph{speaker} and a \emph{listener}. 
Each game starts with 10 tangrams as the \emph{context}, with 3--5 tangrams designated as \emph{targets}. 
The target designations are revealed to the speaker but hidden to the listener. The goal is to select all targets without selecting any non-targets.
The speaker can only communicate with the listener through a sequence of utterances, and only the listener can take selection and deselection actions.
The interaction starts with a speaker turn. Turns alternate between speaker and listener, with a maximum of 20 turns. 
In each speaker turn, they type an utterance to send to the listener. 
Speaker turns are limited to 25 seconds. 
In each listener turn, they have 45 seconds to select or deselect images as instructed to by the speaker. 
The game concludes when the listener selects only and all targets, or when the partners run out of turns. 
\aautoref{mturk-details} shows screenshots of the interface. 

\paragraph{Context Construction}
We follow \citet{gul2024CoGen} and construct game contexts using 1{,}013 tangram images from \kilogram~\cite{ji2022Abstract}. We group tangrams randomly into two splits: development split (101 tangrams) and main split (912 tangrams). The development split is exclusively used for seeding the initial listener policy $\policy_{\param_0}$. All human-bot interactions are constructed from the main split, i.e., tangrams that the seed policy $\policy_{\param_0}$ has never seen before. We construct all games with 3--5 target tangrams. More targets are generally harder, given the same maximum number of turns per interaction.

\subsection{Human Evaluation Design}\label{human-eval-design}

Automatically evaluating turn-level policy performance is hard, because we have no ground truth (i.e., the selection and deselection actions intended by the speaker in each turn) to compare against. Similarly, we have no ground truth to systematically assess the feedback decoder quality. We conduct human evaluation surveys to address these problems. We annotate a subset of \bp interactions, roughly 120 interactions or 1{,}000 turns per system-turn.

We show human annotators a complete interaction turn by turn, without revealing the underlying targets. For each turn, the annotation consists of two phases:

\begin{enumerate}
    \item Ground-truth: we show context, currently selected tangrams, and instruction given by the speaker. We ask the annotator to annotate the listener action. The annotator action  $\action^*$ is considered as ground truth action for this turn. We use these labels for turn-level evaluation. After the action annotation, we reveal the action $\hat\action$ actually taken by the listener (i.e., the model) during the interaction. 
    \item Satisfaction: we present the follow-up utterance. We ask the annotator to rate if the speaker is satisfied with the listener's action, based on the follow-up utterance. They choose one of the following options:
    \begin{enumerate}[label=\alph*.]
        \item Yes.
        \item Yes, even though the listener did not perform all required selections/deselections.
        \item Yes, even though the listener made incorrect selections/deselections.
        \item No.
    \end{enumerate}
    The third option accounts for the listener  accidentally selecting a target tangram not intended by the speaker, but the speaker choosing to move on without correction or even validating the selection. We treat these labels as ground truth for evaluating feedback decoders.
\end{enumerate}

We annotate 5\% of long-term human-bot interactions annotations by three different annotators, to estimate how reliable the annotations are. 
We observe 85\% agreement on the correctness (whether $\hat\action = \action^*$) on ground truth stage,\footnote{The percentage of cases where all annotators agree that the bot did right or wrong.} and 65\% agreement on the ground-truth action $\action^*$ across workers.\footnote{The percentage of cases where all three annotators provided exactly the same set of actions.}
For satisfaction annotation, we observe 93\% agreement rate, illustrating the simplicity of extracting the signal that drives our learning process.

\subsection{MTurk Details}\label{mturk-details}

\paragraph{Worker Recruitment}
We follow \citeauthor{gul2024CoGen}'s (\citeyear{gul2024CoGen}) worker recruitment recipe. 
We require workers to have a minimum 98\% approval rate, at least 1{,}000 approved HITs (Human Intelligence Task), and be located in English-majority locales. All workers must watch a video tutorial and pass a quiz before gaining qualification to work on \multiref interactions. They must read a thorough guideline and pass another quiz before being granted access to human evaluation surveys. We recruit 33 expert workers to interact with LLMs in the main study and annotate by completing surveys after the main study. They are required to accept a consent form detailing how MTurk worker IDs are encrypted, how the collected data would be published, and risks of participating in this study.
This study is exempted from \eat{Cornell's}Institutional Review Board.

\paragraph{Payment}
We pay workers \$0.81 USD per \multiref game, and a bonus if the game is successful. Overall the estimated hourly wage is \$13.00 USD, and closer to \$23.00 USD by the end of the continual study when the LLM is fairly good at the game. On average a human-bot game takes under 2 minutes. We pay workers \$0.06 USD per turn for human evaluation surveys, or \$0.08 USD if the turn annotation involves error modes. The estimated hourly wage is \$16.00 USD for human evaluation surveys. On average it takes under 2.5 minutes to annotate one game. We set the payment scheme through pilot studies and aim for a \$15.00 USD hourly wage.

\begin{figure*}
    \centering
    \includegraphics[clip,trim=0 1in 2in 0, width=0.99\textwidth]{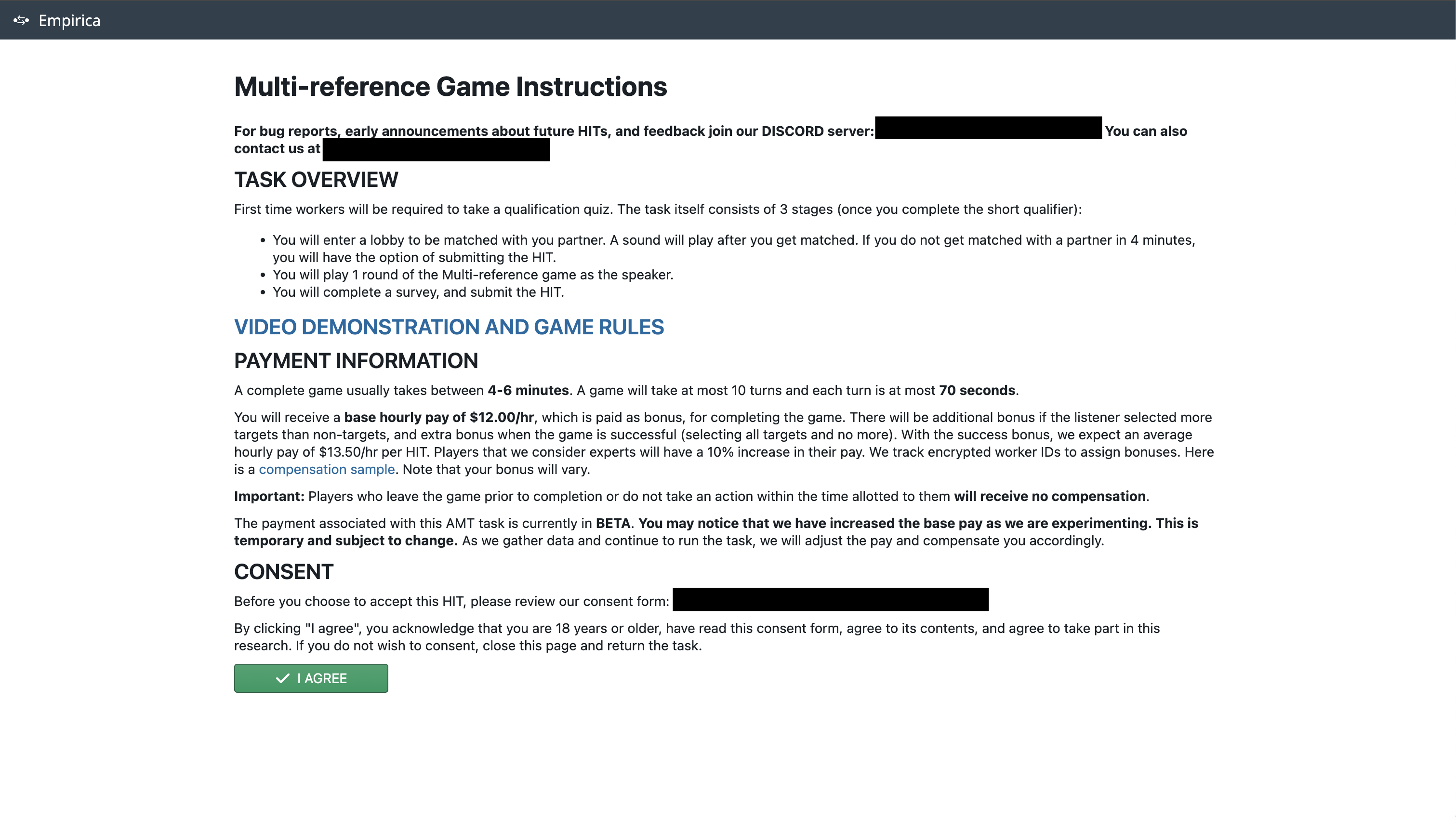}
    \caption{The \multiref instruction page for workers including compensation details and game rules.}
    \label{fig:interface-screenshot-preview}
\end{figure*}

\begin{figure*}
    \centering
    \includegraphics[clip,trim=0 2in 2in 0, width=0.99\textwidth]{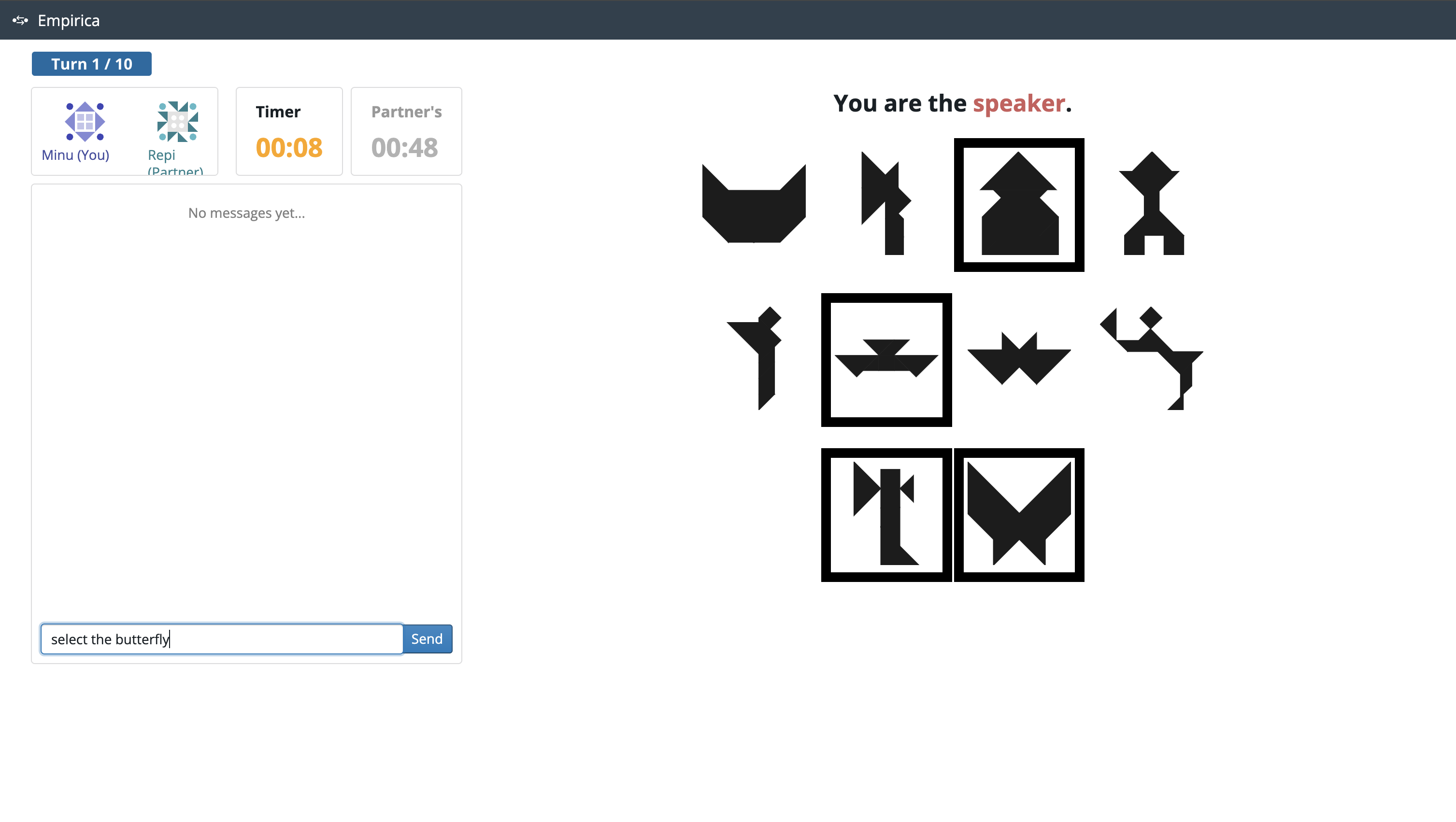}
    \caption{The \multiref interface for the speaker in turn 1. Predefined targets are revealed to the speaker in black boxes.}
    \label{fig:interface-screenshot-speaker1}
\end{figure*}

\begin{figure*}
    \centering
    \includegraphics[clip,trim=0 2in 2in 0, width=0.99\textwidth]{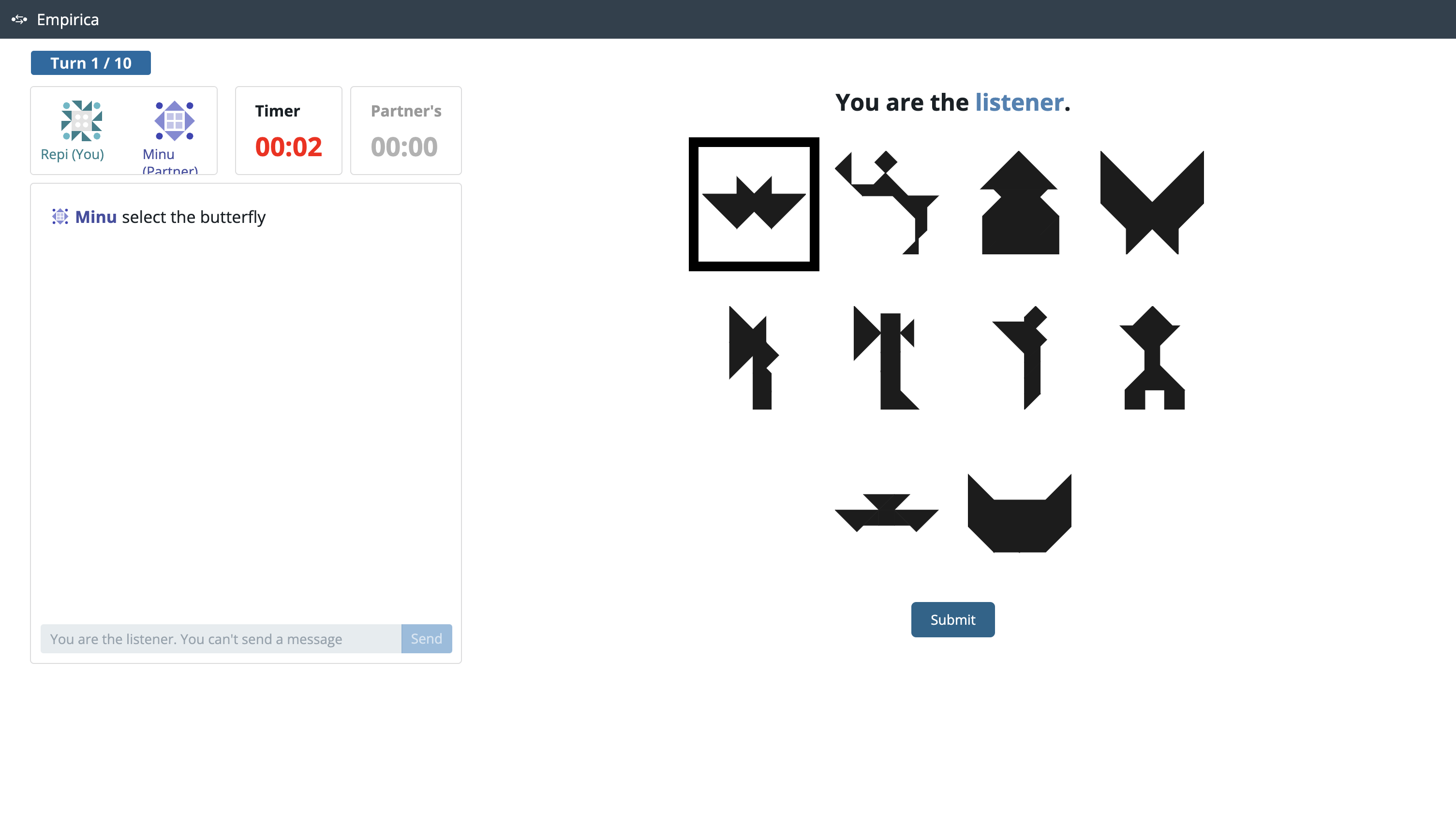}
    \caption{The \multiref interface for the listener in turn 2, following the speaker turn in \autoref{fig:interface-screenshot-speaker1}. Targets are hidden for the listener, and the context tangrams are in a different order. Here the listener has selected a tangram given the instruction \nlstring{select the butterfly}.}
    \label{fig:interface-screenshot-listener1}
\end{figure*}

\begin{figure*}
    \centering
    \includegraphics[clip,trim=0 1in 2in 0, width=0.99\textwidth]{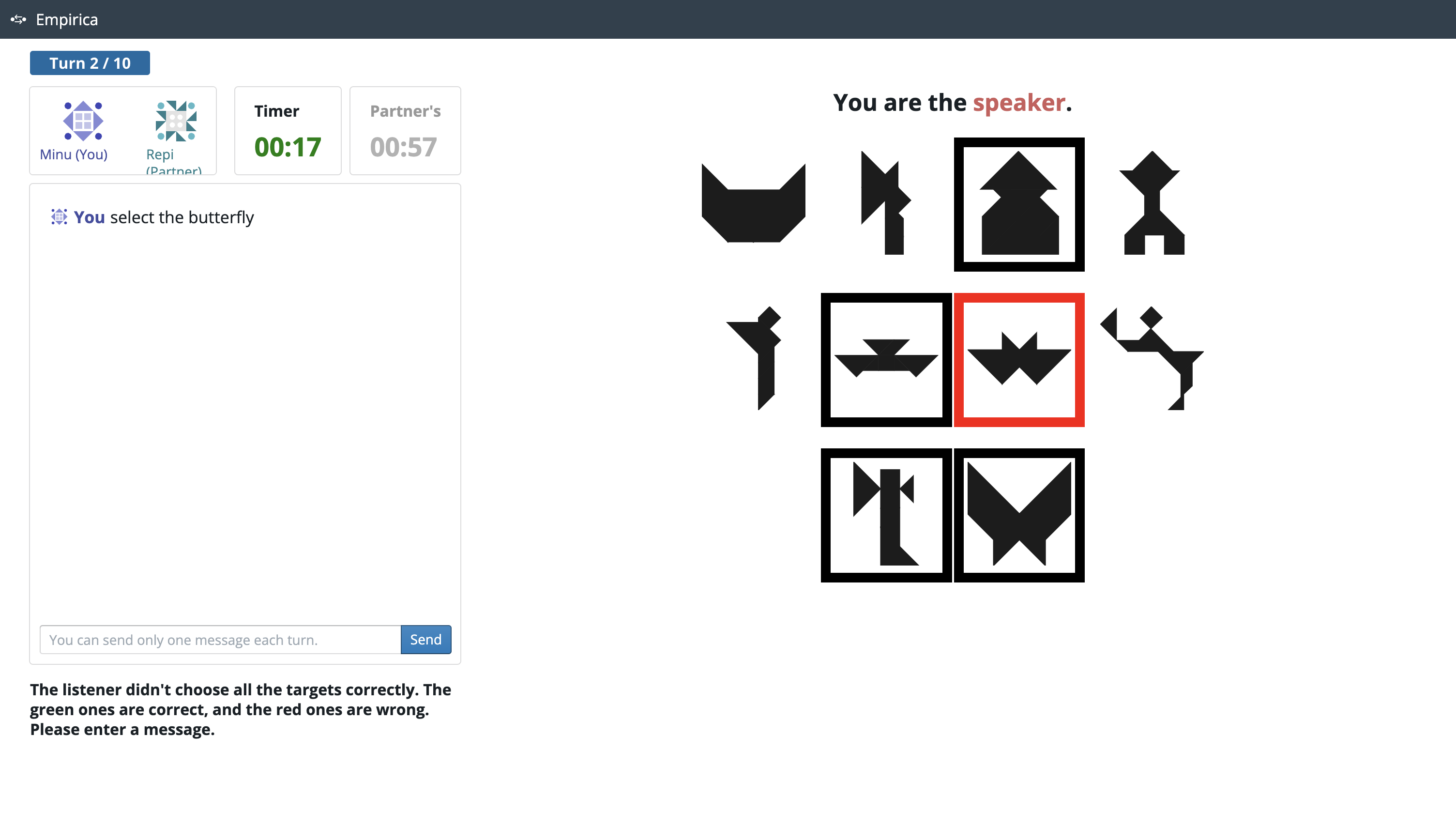}
    \caption{The \multiref interface for the speaker in turn 3, following the listener turn in \autoref{fig:interface-screenshot-listener1} The listener selected a non-target tangram, shown in red to the speaker.}
    \label{fig:interface-screenshot-speaker2}
\end{figure*}

\paragraph{Interface and serving}
We implement \multiref using Empirica~\citep{almaatouq2021Empirica} and on top of the code base of \citet{gul2024CoGen}. The speaker has 25 seconds to type into a chat box each turn and hit Enter or submit, and the listener has 45 seconds to click on the tangrams to select or to deselect. The game ends if one party idles for one turn, and the party idling is not compensated. We serve on an EC2 instance. We serve LLM policies with the Ray framework~\citep{moritz2018Ray}. We walk through the first turns of a sample interaction in 
\autoref{fig:interface-screenshot-preview},
\autoref{fig:interface-screenshot-speaker1}, \autoref{fig:interface-screenshot-listener1}, and \autoref{fig:interface-screenshot-speaker2}.

\subsection{Artifacts Used and Data Release}

\paragraph{Licenses}
The scientific artifacts we used, including the \Idefics model~\citep{laurencon2024What} and the model serving framework Ray~\citep{moritz2018Ray}, have open licenses (the Apache 2.0 License). We will release our model checkpoints and human-model interactions under Apache 2.0 as well.

\paragraph{Privacy Precaution}
We store encrypted MTurk worker ID via md5 hashing in our database. No other personal identifiable information is collected or released.

\section{Learning Details} \label{learning-details}

\subsection{Interaction Representation}

We encode the context $\context$ as in \autoref{fig:policy-prompt}. 
We standardize action representation by ordering actions, for example, always produce \code{Select A C} rather than \code{Select C A}. We shuffle the context images during training as the order of context tangrams should not have any impact on the interaction logic.

\begin{figure}
    \centering
    \begin{tcolorbox}[width=0.48\textwidth, boxsep=0pt, left=4pt, right=4pt, top=4pt, bottom=4pt, sharp corners]
        System: \tkimg A \tkimg B \tkimg C \tkimg D \tkimg E \tkimg F \tkimg G \tkimg H \tkimg I \tkimg J \boxcomment{Context tangrams tagged by code names}\\        
        System: none is selected \boxcomment{Turn 1 state} \\
        User: Select the letter S with the large triangle back side \boxcomment{Turn 1 utterance}\\
        Assistant: Select J \boxcomment{Turn 1 action}\\
        System: J currently selected \\
        User: unselect that one and try again \\
        Assistant: Deselect J select A \\
        System: A currently selected \\
        User: almost, unselect that one and try one more time \\
        Assistant: Deselect A select G \\
        System: G currently selected \\
        User: correct. Now select the rabbit facing right \\
        Assistant: Select F \\
        System: F G currently selected \\
        User: no that is a bird, try again \\
        Assistant: Deselect F select H \\
        System: G H currently selected \\
        User: no that is another bird, try again \\
        Assistant: \textbf{Deselect H select F} \boxcomment{Action}
    \end{tcolorbox}
    \caption{Policy prompt example with a model predicted \textbf{action} and additional \textcolor{c0}{\emph{comments}} for readability.}
    \label{fig:policy-prompt}
\end{figure}

\subsection{Policy Initialization}\label{learning-details-seeding}

We seed the initial policy $\policy_0$  by fine-tuning the model on a small dataset of 90 successful turns $\dataset_0$, where both the speaker and the listener are humans.
We also experiment with prompting to initialize the policy. We find early that few-shot prompting yields a random policy at best, likely because reasoning with abstract shapes such as tangrams is visually out-of-distribution for the model.

There is a significant distribution shift between human-human interactions, and human-policy interactions, especially early on when the model performs poorly. 
In practice, two major differences are the length of interactions and the prevalence of deselection instructions, which are rare in human-human interactions. 
We address the deselection issue with data augmentation. 
We synthetically generate turns where the speaker asks for deselections, and the listener complies. 
We augment the data with these at a ratio of 1:12 to the existing data. 
This helps the LLM policy learn to deselect and recover from mistakes. This augmentation is only used for $\dataset_0$ and such distribution shift is not present in later rounds, when learning from actual human-bot interactions.

We validate our design online with 30 main-split human-bot pilot interactions, or offline with a validation set of 344 successful main-split human-human turns.

\begin{figure}
    \centering
    \begin{tcolorbox}[width=0.48\textwidth, boxsep=0pt, left=4pt, right=4pt, top=4pt, bottom=4pt, sharp corners]
        [Previous turns omitted]\\
        System: none is selected \boxcomment{Previous turns}\\
        Speaker: Man in a hat\\
        Listener: Select A\\
        System: A currently selected \boxcomment{Augmented state}\\
        Speaker: Wrong, undo what you selected  \boxcomment{Augmented utterance}\\
        Listener: \textbf{Deselect A} \boxcomment{Augmented action}
    \end{tcolorbox}
    \caption{An example of deselection augmentation with augmented \textbf{action} and \textcolor{c0}{\emph{comments}}.}
\end{figure}

\subsection{Hyperparameters and Other Implementation Details} \label{learning-details-hyperparameters}

We use the instruction-tuned \Idefics model for all policies. We fine-tune with LoRA adapters~\citep{hu2021LoRAa} ($\alpha$=$r$=8, dropout=0.1) due to compute constraints. \aautoref{lora-enhanced} provides more LoRA details.
We train each model with a single GPU, RTX A6000, NVIDIA A100 40GB or 80GB. The time to train ranges between 2--24 hours, longer in later rounds as more data accumulates. For stopping criteria,
we pick checkpoints by highest accuracy (exact match) among three seeds on a held-out validation set of 344 turns $D_{\text{val}}^{\hh}$. The validation set is curated from 92 human-human games in the main split of tangrams. We summarize hyperparameters in \autoref{tab:hyperparameters}.

\begin{table*}[h]
    \centering
    \begin{tabular}{l c ccc}
        \toprule
        Hyperparameter & Search Space & FFT & RL & KTO \\
        \midrule
        Optimizer &  & AdamW & AdamW & RMSProp \\
        Learning rate & \{1e-6, 1e-5, 1e-4, 2e-4\} & 1e-4 & 1e-4 & 1e-5 \\
        Learning rate decay & \{no, cosine, linear\} & cosine & cosine & no \\
        Epochs & \{5, 10, 20, 40\} & 20 & 20 & 20 \\
        Warm-up steps & \{0, 10, 50\} & 10 & 10 & 10 \\
        Weight decay & \{0, 0.01, 0.1\} & 0.01 & 0.01 & 0.01 \\
        Effective batch size & \{16, 32, 48, 64, 128\} & 64 & 64 & 64 \\
        Entropy weight & \{0, 0.01, 0.5, 0.1\} &  0.01 & 0.01 & 0.1 \\
        $\beta_\text{KTO}$ & \{0.01, 0.1, 0.5\} &   &   & 0.5 \\
        Temperature &  & 1 & 1 & 1 \\
        \bottomrule
    \end{tabular}
    \caption{Hyperparameter settings.}
    \label{tab:hyperparameters}
\end{table*}

\paragraph{Data imbalance} 
The decoded feedback is imbalanced, with more negative examples than positive examples (3:1 to 2:1), especially at early rounds of continual learning. We address this by weighting the loss by the absolute value of the reward, i.e., $-0.1$ for RL or $\lambda_d$ and $\lambda_u$ for KTO, and by downsampling negative examples per batch, such that the number of positive examples and negative examples is roughly 5:4.

\paragraph{KTO stability}
Deviation from the original KTO implementation by higher learning rate, higher $\beta$, more epochs, produces better results empirically on the validation set in pilot and round $\round=1$. However, in round $\round=2$, \bk policy starts to degenerate by producing nonsensical actions such as \code{Deselect A select A B} or \code{Deselect select select}. We attempt to mitigate this issue during training round $\round=3$ by switching from weighting $\lambda_d=4$ and $\lambda_u=1$ as recommended in \citet{ethayarajh2024KTO} to $\lambda_d=\lambda_u=1$, plus downsampling negative examples. We also introduce regex-based constrained decoding to prevent nonsensical actions for \bk and \tk policies in round $\round=3$. Despite that, the KTO group performs worse in live interactions (\autoref{fig:task-performance}). 
We suspect KTO is more challenging to optimize for iterative continual learning, but we suspect further tuning (with higher computational costs) can reduce or even eliminate these issues.

\subsection{Evaluation Metrics}\label{learning-details-metrics}

\paragraph{Interaction-\relax level metrics}

Interaction performance and statistics are computed automatically from live deployment interactions. They do not require further annotation. 

\begin{enumerate}
    \item \textbf{Success rate} = \# successful interactions / \# all interactions. An interaction is successful if the listener selects all and only targets before running out of 10 turns. This is the primary metric we use to evaluate the performance of the LLM policy.
    \item \textbf{\# Turns per interaction}. This is a measure of collaborative efficiency.
\end{enumerate}

\paragraph{Turn-\relax level metrics with reference to human annotation}

We compute turn-level metrics either with respect to \hh games where we consider human listener action as ground truth (e.g., $\dataset_\text{val}^\hh$), or with respect to \bp games where we consider actions $\action^*$ annotated in post-hoc surveys as ground truth. 
When computed with live interactions, these metrics are biased towards longer or failed interactions because they have more turns than successful interactions.

\begin{enumerate}
    \item \textbf{Exact match} = \# exact match / \# all turns. An exact match is when the action taken by the policy matches exactly the action labeled/taken by human listeners ($\hat\action = \action^*$).
    \item \textbf{Similarity} = $\text{Sim}(\hat\action, \action^*)$ is a composite metric. Let $f(p, q): \mathcal{I} \times \mathcal{I} \rightarrow \reals$ be a function  that evaluates the similarity between two images $p, q \in \mathcal{I}$. Let the action taken by policy be $\hat\action = 
    \{ \hat p_1, \hat p_2, ..., \hat p_{\hat{n}}, \hat q_1, \hat q_2, ..., \hat q_{\hat{m}} \}$ where $p$ are the selected tangrams and $q$ are the deselected tangrams. Denote the ground truth actions as $\action^* = \{ p_1^*, p_2^*, ..., p_{n^*}^*, q_1^*, q_2^*, ..., q_{m^*}^* \}$. The similarity between two actions is defined as:

    \begin{align}
    \text{Sim}(\hat\action, \action^*) &= \frac{1}{\hat{n}n^* + \hat{m}m^*} \Bigg(  
    \sum_{i=1}^{\hat{n}} \sum_{j=1}^{n^*} f(\hat p_i, p_j^*)  \notag \\
    &\quad + \sum_{i=1}^{\hat{m}} \sum_{j=1}^{m^*} f(\hat q_i, q_j^*) \Bigg).
\end{align}

    If only one of $\hat n$ and $n^*$ is zero, we rewrite $\Sigma_{i=1}^{\hat{n}} \Sigma_{j=1}^{n^*} f(\hat p_i, p_j^*)$ with $-\max(\hat n, n^*)$, and $\hat nn^*$ in the denominator with $\max(\hat n, n^*)$, intuitively assigning -1 for each missed selection. This edge case is similarly treated for $\hat m$, $m^*$ and deselection. We compute similarities using embeddings from the tangram fine-tuned CLIP model of \citet{ji2022Abstract}
.    
    \item \textbf{Positive feedback} = \# turns receiving positive feedback / \# all turns. An action receives positive feedback if speaker is satisfied with the listener's action in the follow-up interaction. This is labeled in human evaluation survey.
\end{enumerate}

Intuitively, a click is approximately accurate if it selects a target or deselects a non-target. We compute this for all clicks from all interactions in a round for all systems in \autoref{fig:task-performance}.

\paragraph{Corpus-\relax level metrics}

We analyze speaker instructions per system-round. The keywords used to generate the analysis in \autoref{fig:language-analysis} are:

\begin{enumerate}
    \item \textbf{\# Reset} = occurrences of phrases in \{\nlstring{reset, restart, from scratch, all over, start over, deselect everything, deselect all, remove everything, remove all, clear everything, clear all, unselect everything, unselect all, drop everything, drop all} \}
    \item \textbf{\# Try again} = occurrences of phrases in \{\nlstring{try again, try one more time, the other one} \}
\end{enumerate}

\section{Cumulative Number of Interactions Observed}\label{cumsum}

\begin{figure}[h]
    \centering
    \includegraphics[width=0.48\textwidth]{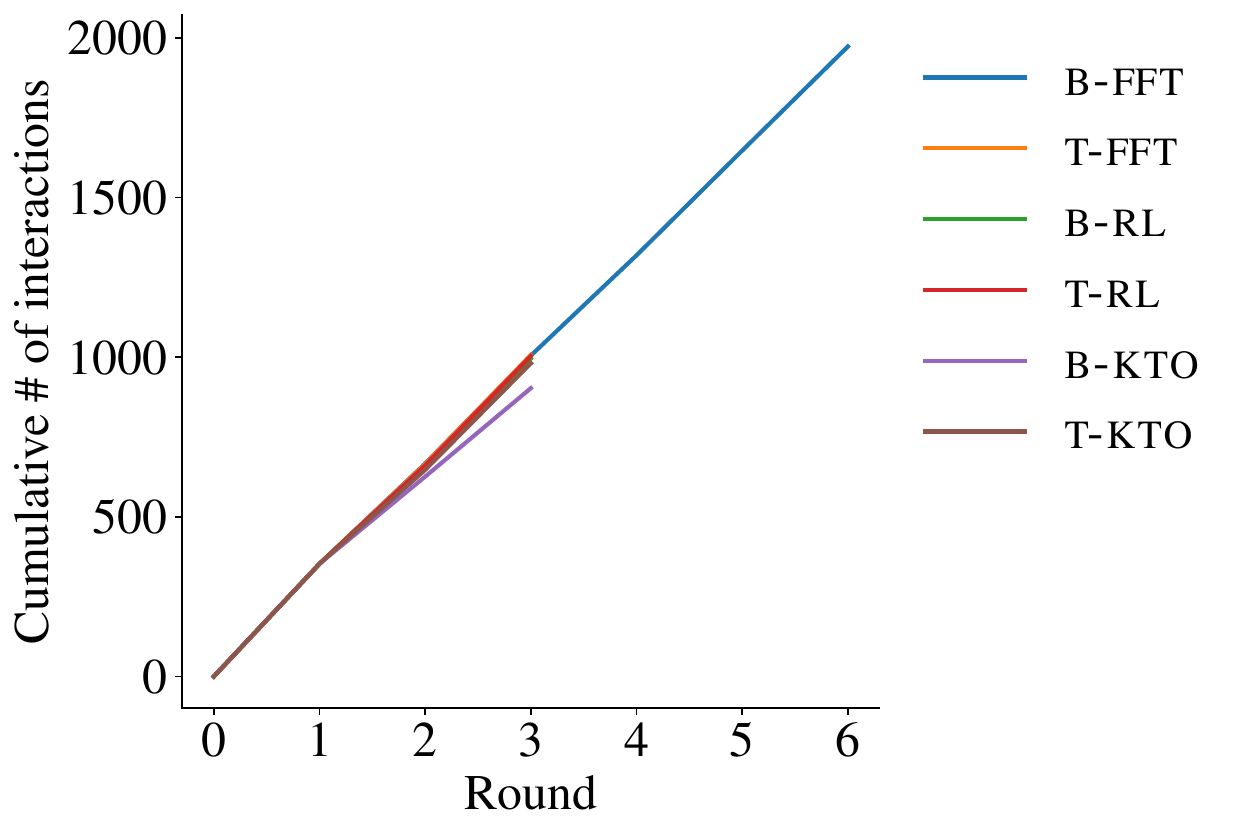}
    \caption{Cumulative number of human-bot interactions used to train the policy each round.}
    \label{fig:cumsum}
\end{figure}

The main text includes results by round. 
We collect roughly 330 interactions per policy per round. Due to the uncertainty of live data collection, we do not always hit this exact number for each variant and round. 
\autoref{fig:cumsum} shows the cumulative number of human-bot interaction seen by a policy variant in each round. 

\section{Additional Enhanced LoRA Launch} \label{lora-enhanced}

\begin{figure}[h]
    \centering
    \includegraphics[width=0.4\textwidth]{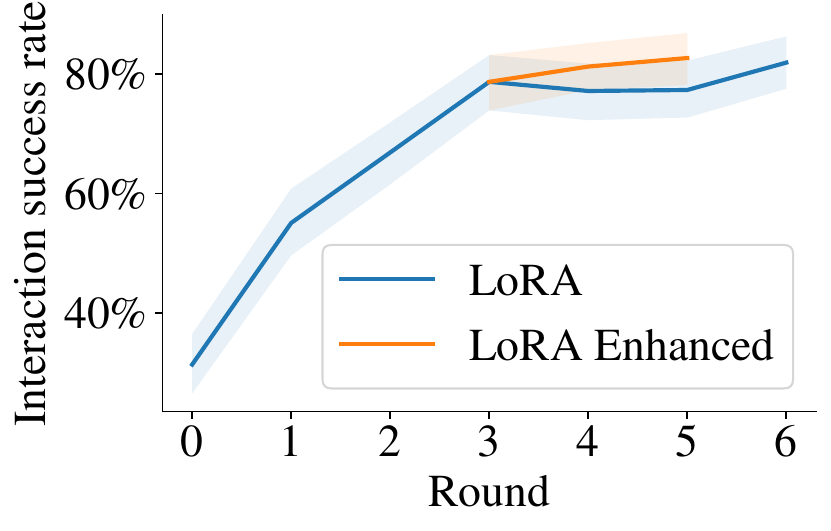}
    \caption{Success rate of \bp with additional LoRA adapters in round 4 and 5.}
    \label{fig:lora-enhanced}
\end{figure}

We suspect the plateau of \bp in \autoref{fig:task-performance} is partially due to the limited expressivity of LoRA adapters we used. 
We test this hypothesis by deploying round $\round=4$ and $\round=5$ again with enhanced LoRA adapters. We use the same hyperparameters as in \autoref{learning-details-hyperparameters} except for the additional adapters. 
The original adapter placement is on the text model, the modality projector, and the perceiver resampler. 
Adapters include the down projection layers, the gate projection layers, the up projection layers, and the key/query/value projection layers. 
In comparison, the enhanced launch adds adapters on the vision model, including the out projection, the first and the second fully connected layers, besides the projection layers on text models. 
\autoref{fig:lora-enhanced} shows the results from this complementary deployment. 
The enhanced LoRA adapters yield a small improvement in interaction success rate compared to the original launch, yet the overall slowdown is evident. This suggests LoRA expressivity has some effect, but other effects are also limiting the LLM policy from continuing its earlier improvement trends.

\section{Detailed Results} \label{tab-results}

We present numerical results of metrics for interaction level performance in
\autoref{fig:interaction} (\autoref{tab:interaction-success}, \autoref{tab:num-turns}, \autoref{tab:click-accuracy}), 
human evaluation performance in \autoref{fig:human-annotations} (\autoref{tab:human-eval-turn}, 
and language analysis in \autoref{fig:language-analysis} (\autoref{tab:ling-vocab}, \autoref{tab:ling-len}, \autoref{tab:ling-reset}, \autoref{tab:ling-lazy}).

\begin{table}
    \centering 
    \resizebox{\columnwidth}{!}{
    \begin{tabular}{lccccccc} 
        \toprule
        Round & 0 & 1 & 2 & 3 & 4 & 5 & 6\\
        \midrule
        \bp & 31.4 & 55.1 & \textbf{66.9} & \textbf{78.7} & 77.1 & 77.3 & 81.9 \\ 
        \tp & 31.4 & 55.8 & 67.8 & 74.0 & - & - & - \\
        \ba & 31.4 & 50.0 & 64.4 & 70.7 & - & - & - \\
        \ta & 31.4 & \textbf{56.8} & 62.4 & 70.3 & - & - & - \\
        \bk & 31.4 & 45.1 & 52.0 & 46.9 & - & - & - \\
        \tk & 31.4 & 50.0 & 61.7 & 66.1 & - & - & - \\
        \control & 31.4 & - & - & - & - & - & 33.0 \\
        \hh & - & - & - & - & - & - & 100.0 \\
        \bottomrule
    \end{tabular}
    }
    \caption{Interaction task success rate in percentage ($\uparrow$). We collect roughly 330 human-bot games per datapoint, except for \hh where we only collect 50 games. Round 0 is shared among systems, except for \hh. All system are deployed for three rounds, and the top performing one (\bp) is deployed for additional three rounds; preempted or not-applicable rounds are marked with dash (-). We \textbf{bold} the highest task success rate in a round.}
    \label{tab:interaction-success} 
\end{table}

\begin{table}
    \centering
    \resizebox{\columnwidth}{!}{
    \begin{tabular}{lccccccc} 
        \toprule Round & 0 & 1 & 2 & 3 & 4 & 5 & 6 \\ 
        \midrule 
        \bp & 8.87 & 8.16 & \textbf{7.33} & \textbf{6.99} & 6.92 & 6.87 & 6.71 \\ 
        \tp & 8.87 & 7.95 & 7.44 & 7.14 & - & - & - \\ 
        \ba & 8.87 & 8.10 & 7.42 & 7.22 & - & - & - \\ 
        \ta & 8.87 & \textbf{7.94} & 7.60 & 7.24 & - & - & - \\
        \bk & 8.87 & 8.15 & 7.66 & 8.03 & - & - & - \\
        \tk & 8.87 & 8.06 & 7.56 & 7.27 & - & - & - \\
        \control & 8.87 & - & - & - & - & - & 8.75 \\ 
        \hh & - & - & - & - & - & - & 4.61 \\ 
        \bottomrule 
    \end{tabular} 
    }
    \caption{\# turns per interaction ($\downarrow$). Maximum 10 turns. Each game has 3-5 targets and \hh games usually take one turn per target. We \textbf{bold} the fewest \# turns per interaction in a round.}
    \label{tab:num-turns}
\end{table}

\begin{table}
\centering 

\resizebox{\columnwidth}{!}{
\begin{tabular}{lccccccc} 
    \toprule 
    Round & 0 & 1 & 2 & 3 & 4 & 5 & 6 \\
    \midrule 
    \bp & 59.7 & 64.0 & \textbf{67.2} & \textbf{69.9} & 69.8 & 69.5 & 72.2 \\
    \tp & 59.7 & \textbf{65.2} & 67.1 & 68.9 & - & - & - \\
    \ba & 59.7 & 63.8 & 66.6 & 68.8 & - & - & - \\
    \ta & 59.7 & 64.9 & 65.0 & 67.0 & - & - & - \\
    \bk & 59.7 & 60.7 & 61.6 & 58.5 & - & - & - \\
    \tk & 59.7 & 62.1 & 63.2 & 64.0 & - & - & - \\
    \control & 59.7 & - & - & - & - & - & 60.5 \\
    \hh & - & - & - & - & - & - & 89.3 \\
    \bottomrule
\end{tabular}
}

\caption{Click accuracy in percentage ($\uparrow$). We \textbf{bold} the highest click accuracy in a round.}
\label{tab:click-accuracy}
\end{table}

\begin{table}

\setlength{\tabcolsep}{3pt}
\centering 
\resizebox{\columnwidth}{!}{
\begin{tabular}{lccccccc|cc} 
    \toprule 
    Round & 0 & 1 & 2 & 3 & 4 & 5 & 6 & \control & \hh \\
    \midrule 
    Exact match & 30.7& 38.4& 44.8& 47.2& 48.7& 46.7& 52.3& 31.7& 79.1\\
    Pos Feedback & 33.0& 39.2& 43.1& 47.5& 49.1& 49.4& 50.4& 34.6 & 78.4\\
    $\text{Sim}(\hat{a}, a^{*})$ & 19.0 & 34.8 & 42.5 & 46.0 & 47.7 & 43.5 & 51.3 & 19.4 & 83.8 \\ $\text{Sim}(\hat{a}, a^{*})$ \textsc{-FB} & 0.0 & 13.6 & 19.2 & 23.9 & 23.3 & 15.6 & 25.7 & 1.4 & 67.9 \\
    \bottomrule
\end{tabular}
}
\caption{Turn level performance of \bp based on human evaluation, all in percentages ($\uparrow$).}
\label{tab:human-eval-turn} 
\end{table}

\begin{table}
\centering 

\resizebox{\columnwidth}{!}{
\begin{tabular}{lccccccc} 
    \toprule 
    Round & 0 & 1 & 2 & 3 & 4 & 5 & 6 \\
    \midrule 
    \bp & 1458 & 1400 & 1283 & 1179 & 1206 & 1211 & 1307 \\
    \tp & 1458 & 1361 & 1279 & 1206 & - & - & - \\
    \ba & 1458 & 1352 & 1248 & 1187 & - & - & - \\
    \ta & 1458 & 1310 & 1306 & 1164 & - & - & - \\
    \bk & 1458 & 1324 & 1183 & 1238 & - & - & - \\
    \tk & 1458 & 1303 & 1332 & 1184 & - & - & - \\
    \control & 1458 &  - & - & - & - & - & 1311 \\
    \hh & - & - & - & - & - & - & 433 \\
    \bottomrule
\end{tabular}
}

\caption{Vocabulary size of speaker instructions when humans interact with different systems across rounds.}
\label{tab:ling-vocab} 
\end{table}

\begin{table}

\centering
\resizebox{\columnwidth}{!}{
\begin{tabular}{lccccccc}
\toprule
Round & 0 & 1 & 2 & 3 & 4 & 5 & 6 \\
\midrule
\bp & 8.78 & 8.87 & 7.94 & 8.49 & 8.53 & 8.30 & 9.23 \\
\tp & 8.78 & 8.69 & 8.24 & 8.49 & - & - & - \\
\ba & 8.78 & 8.29 & 7.94 & 8.45 & - & - & - \\
\ta & 8.78 & 8.42 & 8.39 & 8.26 & - & - & - \\
\bk & 8.78 & 8.29 & 8.59 & 8.57 & - & - & - \\
\tk & 8.78 & 8.05 & 8.41 & 8.05 & - & - & - \\
\control & 8.78 & - & - & - & - & - & 8.19 \\
\hh & - & - & - & - & - & - & 8.49 \\
\bottomrule
\end{tabular}
}
\caption{Utterance length of speaker instructions when humans interact with different systems across rounds.}
\label{tab:ling-len}
\end{table}

\begin{table}
    \centering
\begin{tabular}{lccccccc}
\toprule
Round & 0 & 1 & 2 & 3 & 4 & 5 & 6 \\
\midrule
\bp & 19 & 11 & 14 & 7 & 6 & 9 & 6 \\
\tp & 19 & 5 & 3 & 2 & - & - & - \\
\ba & 19 & 10 & 17 & 9 & - & - & - \\
\ta & 19 & 3 & 9 & 6 & - & - & - \\
\bk & 19 & 17 & 42 & 47 & - & - & - \\
\tk & 19 & 13 & 21 & 8 & - & - & - \\
\bottomrule
\end{tabular}
    \caption{\# Reset words of different systems across rounds.}
    \label{tab:ling-reset}
\end{table}

\begin{table}
    \centering
\begin{tabular}{lccccccc}
\toprule
Round & 0 & 1 & 2 & 3 & 4 & 5 & 6 \\
\midrule
\bp & 42 & 24 & 14 & 3 & 4 & 4 & 14 \\
\tp & 42 & 15 & 7 & 5 & - & - & - \\
\ba & 42 & 16 & 12 & 4 & - & - & - \\
\ta & 42 & 21 & 15 & 6 & - & - & - \\
\bk & 42 & 20 & 5 & 9 & - & - & - \\
\tk & 42 & 11 & 6 & 6 & - & - & - \\
\bottomrule
\end{tabular}
    \caption{\# Try again words of different systems across rounds.}
    \label{tab:ling-lazy}
\end{table}

\section{Feedback Decoder Details}\label{feedback-decoder-prompt}

\paragraph{Design}
The prompt design is minimal, general, and task-agnostic. We validate the prompt with manual inspection prior to continual learning launch and human surveys. Considering only the most recent two action-utterance turns $\langle \hat\action_{i-1}, \utterance_{i}, \hat\action_{i}, \utterance_{i+1} \rangle$ is sufficient to produce satisfactory decoding results, and more history seems to distract the decoder. We also experimented with numerical reward (i.e., decoding a real number), experimenting with a discretized reward space of \{.0, .1, .5, .9\} . Our experiments show the model is not well calibrated for such decoding.

\paragraph{Feedback Decoder Error and Potential Fix}\label{app:feedback_decoder_errors}
Roughly 15\% of feedback decoder predictions are false negatives, see \autoref{fig:rd-confusion-matrix} top row, and an example in \autoref{fig:fd-false-negative}. 
We handle negatives in different ways in our experiments, but generally negatives examples have less impact than positive ones, so the learner is robust to false negative noise. 
Of course, it does mean that we are losing valuable positive data, and reducing this error rate is an important direction for future work. This can potentially speed up learning further.

\begin{figure}

    \footnotesize
    \centering
    \begin{tcolorbox}[width=0.48\textwidth, boxsep=0pt, left=4pt, right=4pt, top=4pt, bottom=4pt, sharp corners]
        User: Please carefully read the following conversation and answer: Is the very last utterance from the speaker \textcolor{black}{positive or negative} feedback? Often negative feedback include corrections and keywords like no, not, undo, don't, with generally negative sentiment, while positive feedback often includes good, yes, correct, okay, or simply move on to the next stage. \textcolor{black}{Lean towards negative if it sounds neutral.}\\
        (start of the conversation)
        
        Speaker: house
        
        Listener: Select F \boxcomment{Action to focus on}
        
        Speaker: horned roof \boxcomment{Feedback}
        
        (end of the conversation)\\
        Answer a single word, \textcolor{black}{Positive, or Negative}\\
        Assistant: \textbf{Negative}
    \end{tcolorbox}
    \caption{Feedback decoder false-negative example:
    the feedback decoder fails to recognize an implicit positive feedback from the speaker by moving on to the next target. 
    The verbal \textbf{feedback generated by the model} is in bold. Additional \emph{\textcolor{c0}{comments for readability}} are in italics.}
    \label{fig:fd-false-negative}
\end{figure}

\section{Use of AI Assistant}

Copilot is used for code generation assistance. ChatGPT is used for formatting in LaTeX.

\section{Interaction Case Studies}\label{case-studies}

\begin{figure*}
    \centering
    \fbox{
        \includegraphics[width=0.8\linewidth, trim = 0 6cm 3cm 0, clip]{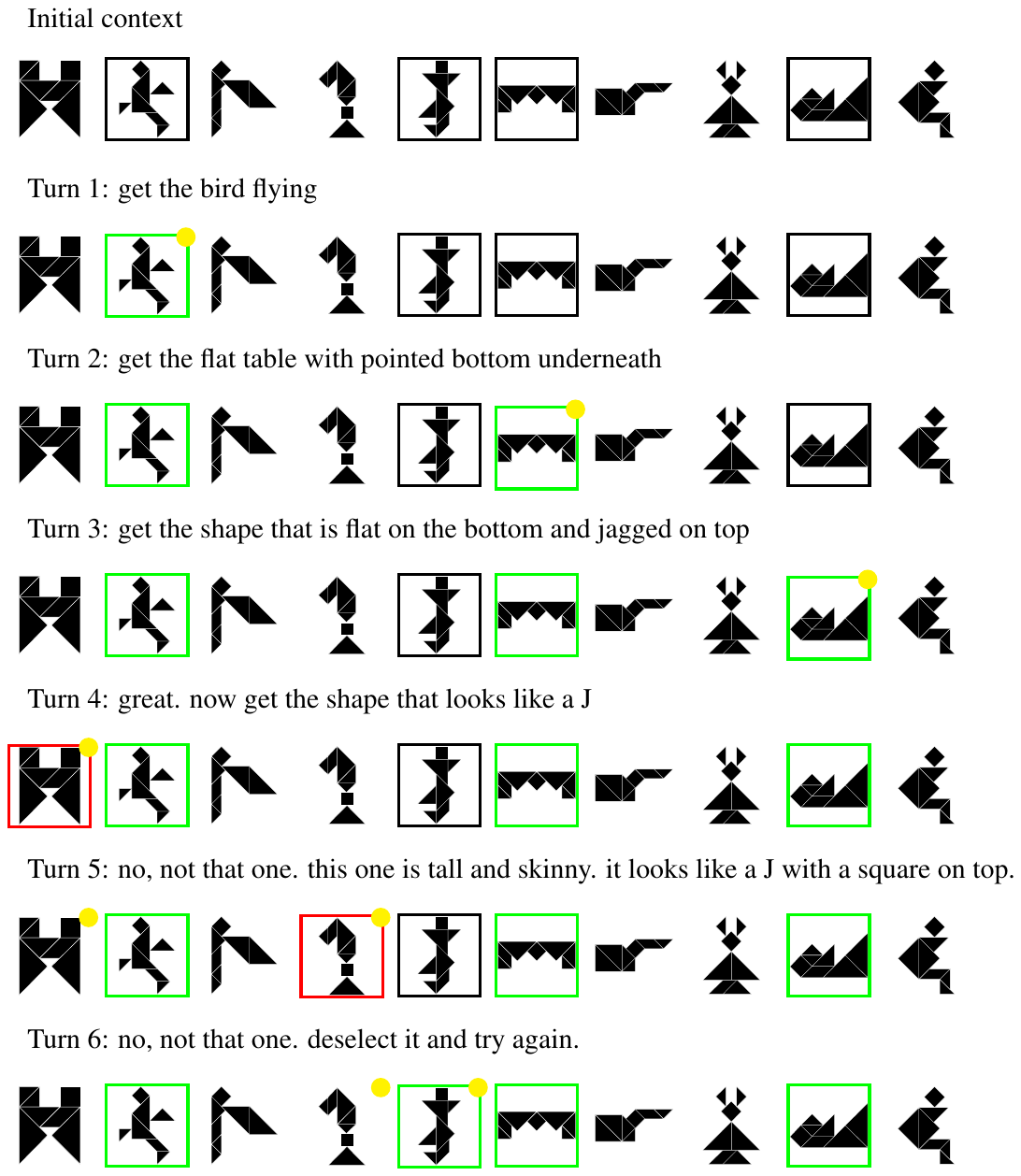}
    }
    \caption{The speaker is left with the last target at Turn 4. Failing, they provide an additional description in Turn 5, and eventually resort to ``try again" without describing the target in Turn 6. The initial turns illustrate how feedback is implied, rather than specified explicitly. The interaction concludes successfully.}
    \label{fig:case-study-try-again}
\end{figure*}

\begin{figure*}
    \centering
    \fbox{
        \includegraphics[width=0.8\linewidth, trim = 0 3cm 3cm 0, clip]{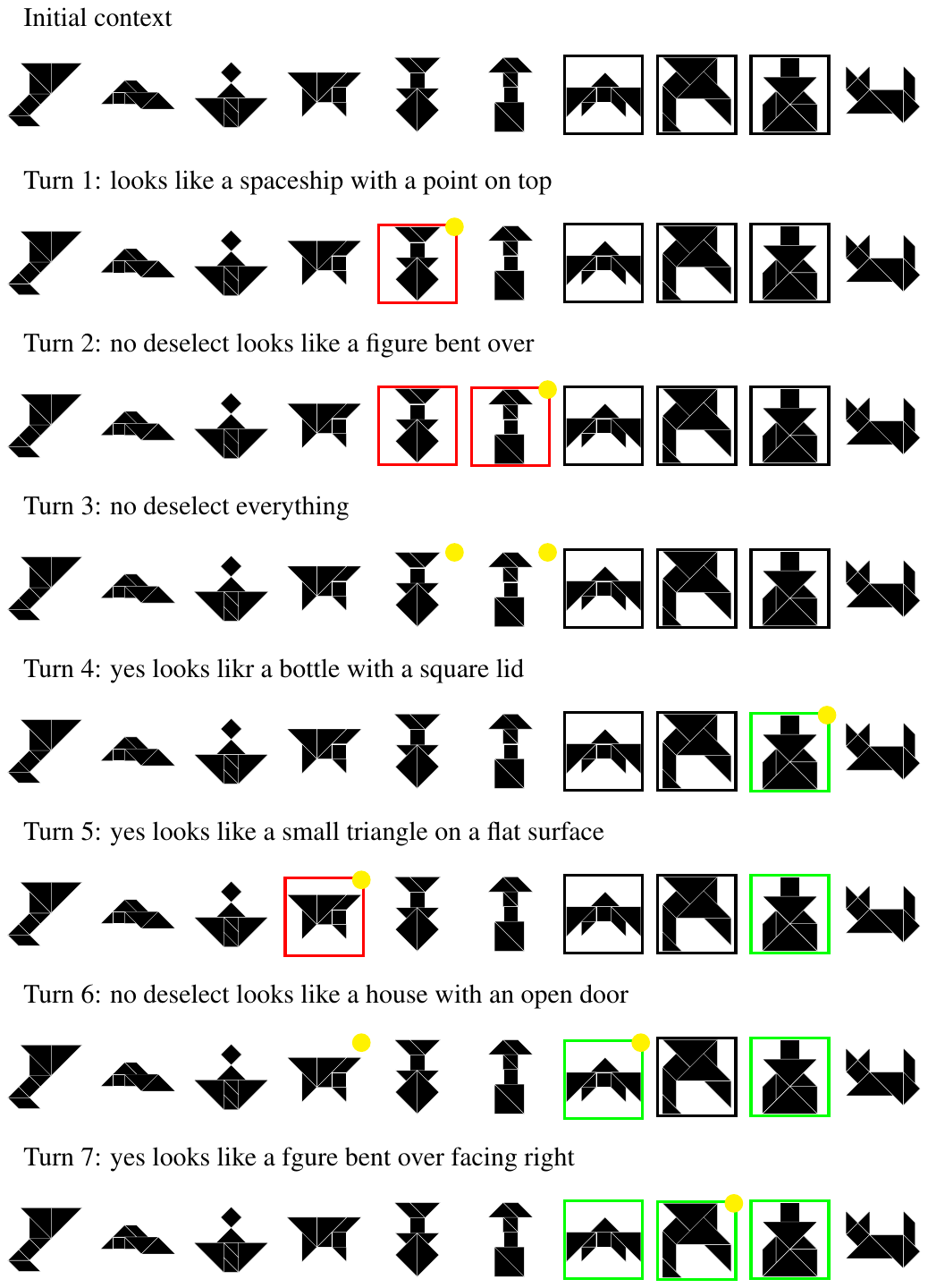}
    }
    \caption{The speaker asks to deselect everything in Turn 3 to reset, an expression of frustration. The interaction concludes successfully.}
    \label{fig:case-study-reset}
\end{figure*}

\begin{figure*}
    \centering
    \fbox{
        \includegraphics[width=0.8\linewidth, trim = 0 3cm 3cm 0, clip]{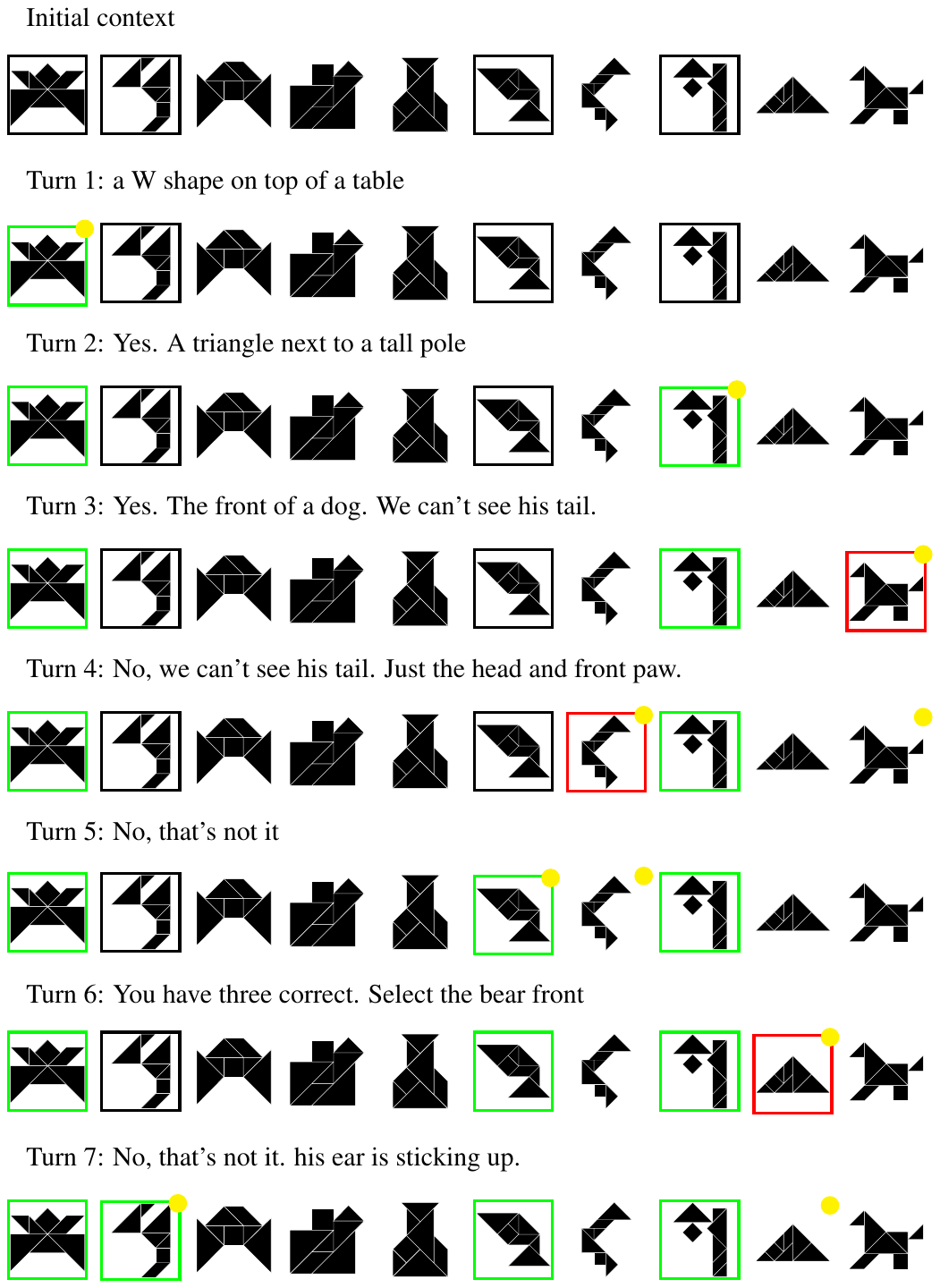}
    }
    \caption{The abstractness and ambiguity of tangrams lend to complex interactions. There are two dogs in the context, and the listener struggles to disambiguate or identify the target. The interaction concludes successfully.}\label{fig:case-study-third}
\end{figure*}

\begin{figure*}
    \centering
    \fbox{
        \includegraphics[width=0.8\linewidth, trim = 0 18cm 3cm 0, clip]{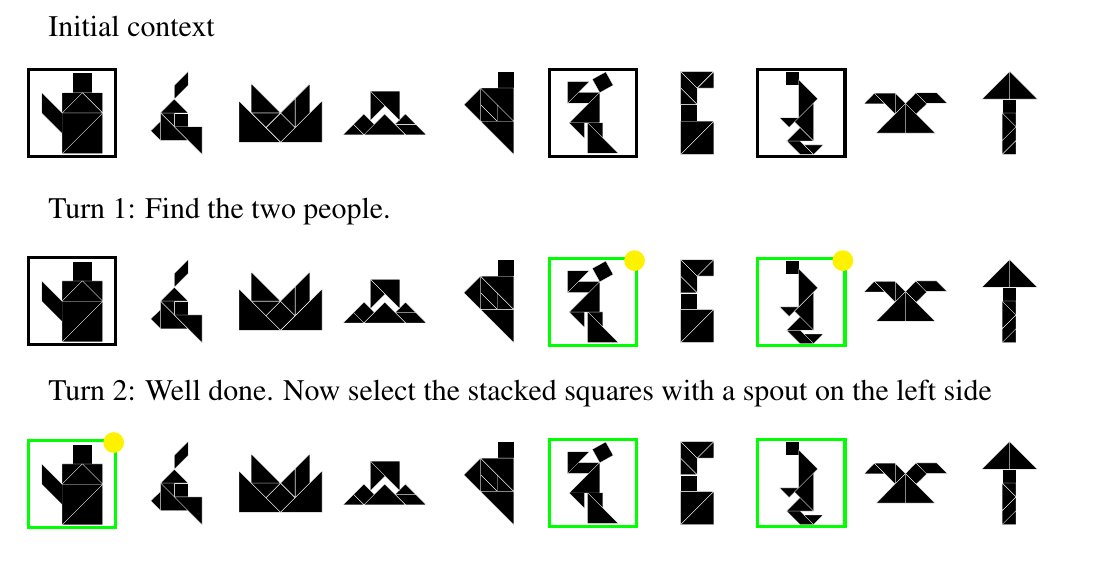}
    }
    \caption{The speaker asks for two targets in Turn 1, exemplifying Grice's Maxims of Quantity - one tries to be as informative as one possibly can, and gives as much information as is needed, and no more~\citep{grice1975logic}. The interaction concludes successfully.}\label{fig:case-study-fourth}
\end{figure*}

Figures \ref{fig:case-study-try-again}--\ref{fig:case-study-fourth} illustrate the diversity of \multiref interaction scenarios. Black borders indicate targets. \textcolor{yellow}{Yellow} dots indicate actions taken by the listener. \textcolor{green}{Green} borders indicate correct selections, while \textcolor{red}{red} borders indicate wrong selection. 

\begin{figure*}
    \centering
    \fbox{
        \includegraphics[width=0.8\linewidth, trim = 0 1cm 0cm 0, clip]{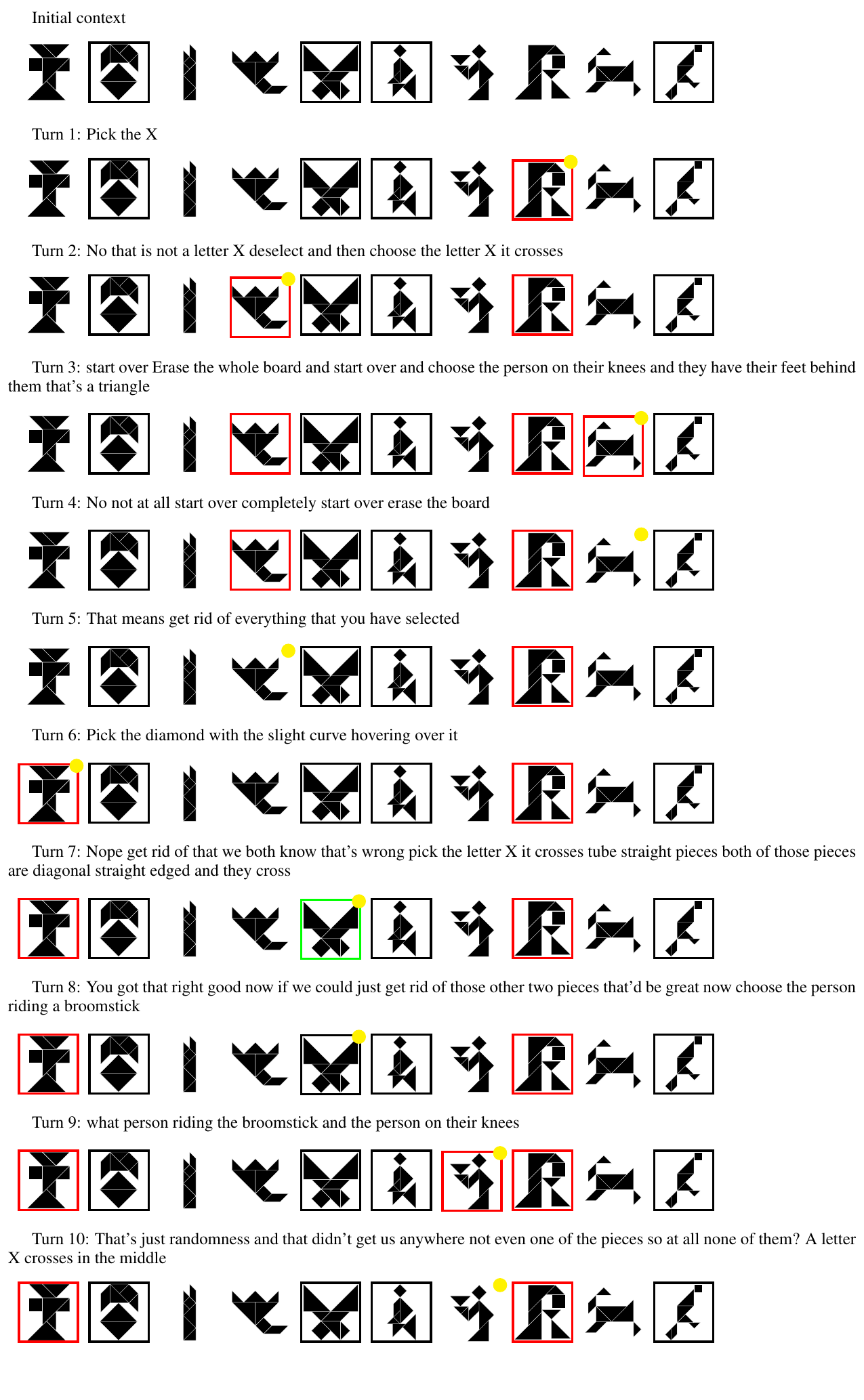}
    }
    \caption{An interaction trace between a human speaker and the initial listener policy $\policy_{\param_0}$. This interaction concludes unsuccessfully.}\label{fig:case-study-triplet-14-initial}
\end{figure*}

\begin{figure*}
    \centering
    \fbox{
        \includegraphics[width=0.8\linewidth, trim = 0 14cm 3cm 0, clip]{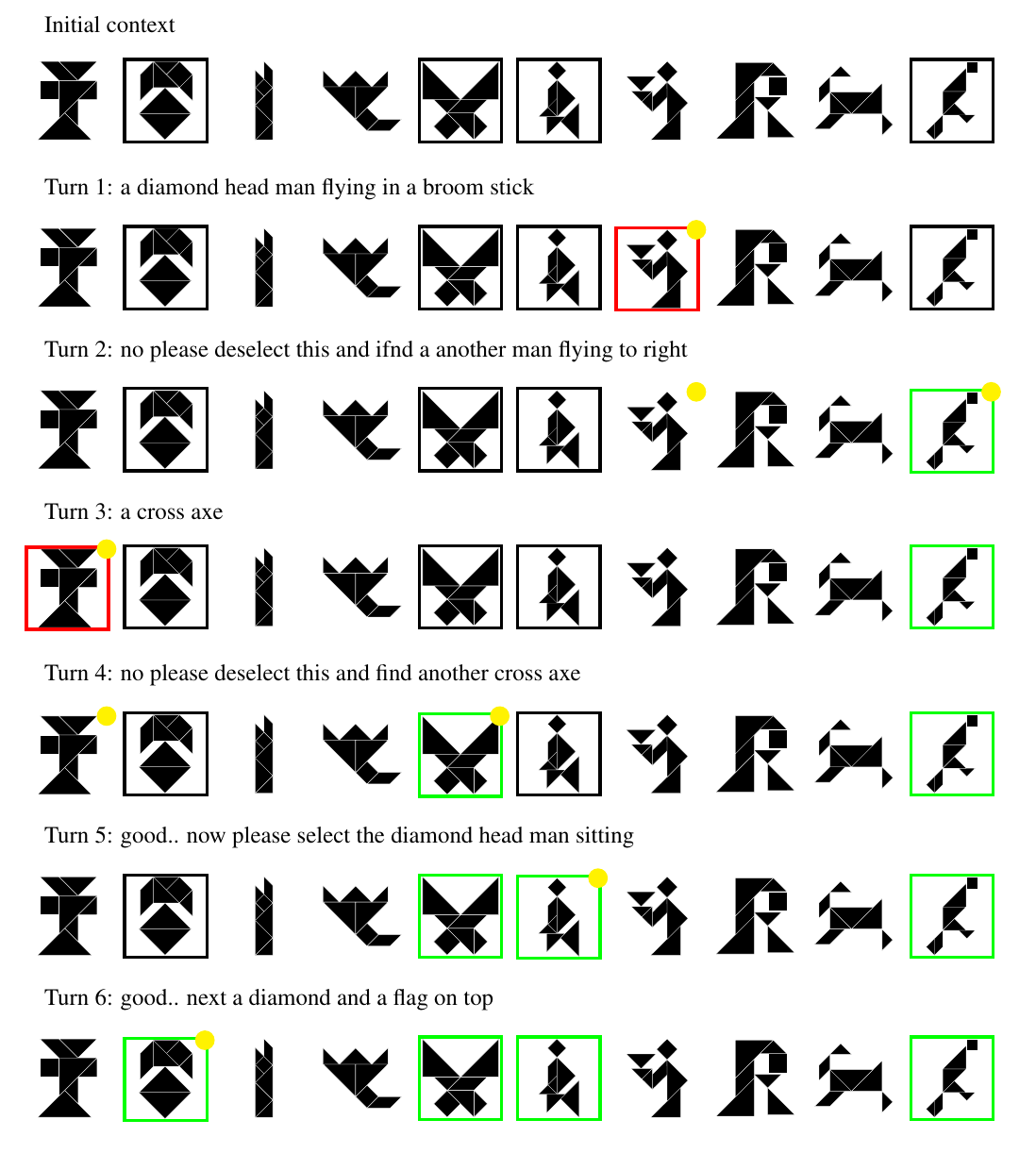}
    }
    \caption{An interaction trace between a human speaker and the final listener policy $\policy_{\param_6}$. This interaction concludes successfully.}\label{fig:case-study-triplet-14-final}
\end{figure*}

\begin{figure*}
    \centering
    \fbox{
        \includegraphics[width=0.8\linewidth, trim = 0 26cm 3cm 0, clip]{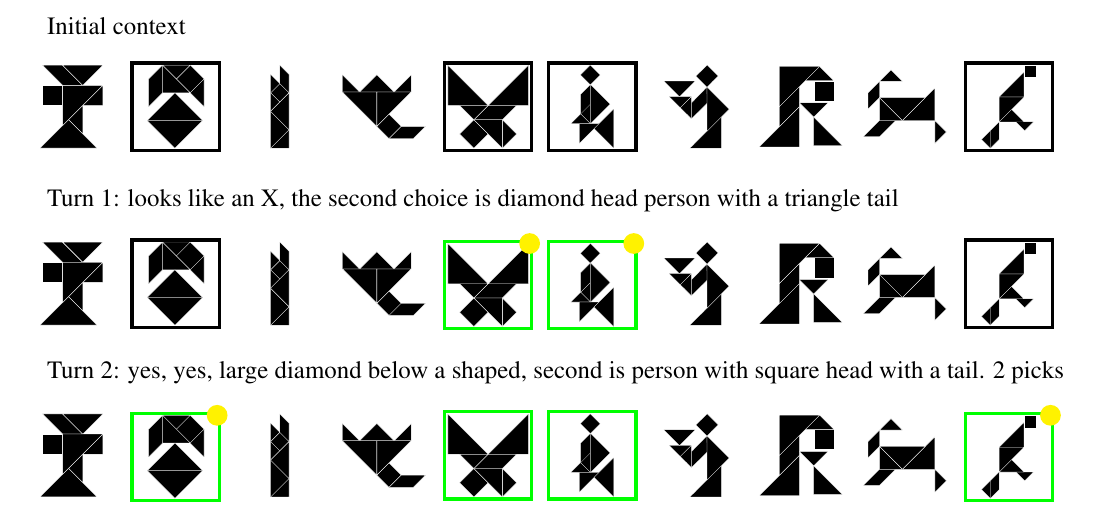}
    }
    \caption{An interaction trace between a human speaker and a human listener. This interaction concludes successfully, even faster than \autoref{fig:case-study-triplet-14-final}}\label{fig:case-study-triplet-14-human-human}
\end{figure*}

We also fix a game context and compare the behavior of the initial $\policy_{\param_0}$ (failed at the game, \autoref{fig:case-study-triplet-14-initial}), the final $\policy_{\param_6}$ (succeeded at the game within 6 turns, \autoref{fig:case-study-triplet-14-final}), and a human listener (succeeded within 2 turns, \autoref{fig:case-study-triplet-14-human-human}).

\end{document}